\pdfoutput=1
\documentclass[11pt]{article}

\usepackage[table]{xcolor}
\usepackage[final]{acl}

\usepackage{times}
\usepackage{latexsym}

\usepackage[T1]{fontenc}
\usepackage[utf8]{inputenc}

\usepackage{microtype}

\usepackage{inconsolata}

\usepackage{graphicx}

\usepackage{ulem}
\usepackage{enumitem}

\usepackage{longtable}
\usepackage{algorithm}
\usepackage{algorithmic}
\usepackage{xspace}
\usepackage{amsmath}
\usepackage{pifont}
\usepackage{cleveref}
\usepackage{multirow}
\usepackage{lipsum}
\usepackage{booktabs}
\usepackage{tcolorbox}
\usepackage{subfigure}  % for subfigures
\usepackage[prependcaption,textsize=scriptsize,color=pink]{todonotes}
\usepackage{amsmath,amsfonts,bm,bbm}

\def\bva{{\bar{\mathbf{a}}}}

\def\eqref#1{Eq.~(\ref{#1})}
\def\1{\mathbf{1}}

\def\va{{\mathbf{a}}}

\def\vc{{\mathbf{c}}}

\def\ve{{\mathbf{e}}}

\def\vg{{\mathbf{g}}}

\def\vq{{\mathbf{q}}}

\def\vs{{\mathbf{s}}}

\DeclareMathAlphabet{\mathsfit}{\encodingdefault}{\sfdefault}{m}{sl}
\SetMathAlphabet{\mathsfit}{bold}{\encodingdefault}{\sfdefault}{bx}{n}

\def\gL{{\mathcal{L}}}

\def\gS{{\mathcal{S}}}

\newcommand{\R}{\mathbb{R}}

\usepackage{amsthm}

\newcommand{\se}{\textsc{SelfElicit}\xspace}
\newcommand{\co}{\textsc{Cot}\xspace}
\newcommand{\fe}{\textsc{FullElicit}\xspace}
\newcommand{\pe}{\textsc{PromptElicit}\xspace}

\newcommand{\cmark}{\textcolor{teal}{\bf\ding{51}}}
\newcommand{\xmark}{\textcolor{red}{\bf\ding{55}}}
\newcommand{\textev}[1]{\textcolor{blue}{\bf #1}}
\definecolor{em}{gray}{0.9}
\newcommand{\cem}{\cellcolor{em}}

\newcommand{\bs}[1]{{\textbf{#1}}}
\newcommand{\sbs}[1]{{\uline{#1}}}
\newcommand{\di}[1]{{ (\small{#1})}}

\renewcommand{\paragraph}[1]{\vspace{0.3em}\noindent\textbf{#1}\hspace{0.5em}}

\title{\se: Your Language Model Secretly Knows \\Where is the Relevant Evidence}

\author{Zhining Liu\textsuperscript{1}, Rana Ali Amjad\textsuperscript{2}, Ravinarayana Adkathimar\textsuperscript{2}, \textbf{Tianxin Wei\textsuperscript{1}, Hanghang Tong\textsuperscript{1}}\\
  \textsuperscript{1}University of Illinois Urbana-Champaign, \textsuperscript{2}Amazon Science\\
  \textsuperscript{1}\texttt{\{liu326, twei10, htong\}@illinois.edu} \textsuperscript{2}\texttt{\{raamjad,adkathi\}@amazon.com}\\
}

\begin{document}
\maketitle
\begin{abstract}
Providing Language Models (LMs) with relevant evidence in the context (either via retrieval or user-provided) can significantly improve their ability to provide better-grounded responses. 
However, recent studies have found that LMs often struggle to fully comprehend and utilize key evidence from the context, especially when it contains noise and irrelevant information—an issue common in real-world scenarios.
To address this, we propose \se, an inference-time approach that helps LMs focus on key contextual evidence through self-guided explicit highlighting.
By leveraging the inherent evidence-finding capabilities of LMs using the attention scores of deeper layers, our method automatically identifies and emphasizes key evidence within the input context, facilitating more accurate and grounded responses \textit{without} additional training or iterative prompting.
We demonstrate that \se brings consistent and significant improvement on multiple evidence-based QA tasks for various LM families while maintaining computational efficiency.
Our code and documentation are available at \url{https://github.com/ZhiningLiu1998/SelfElicit}.
% We show that \se brings a significant QA performance boost (5.0\%-11.7\% average gain) on multiple evidence-based QA tasks across LMs from Llama-3.1, Mistral, and Qwen2.5 families while maintaining computational efficiency.
\end{abstract}

\section{Introduction}\label{sec:int}

Language models have demonstrated remarkable capabilities~\cite{openai2022intro,openai2023gpt,llmchallenges}, yet hallucinations and factual inaccuracies remain a significant limitation~\cite{ji2023survey,wang2023survey}. 
Furnishing the input context with relevant evidence passage is a popular approach to enhance LM's ability to generate correct and well-grounded responses~\cite{incontext_rag,activerag,wei2025instructrag}.
\begin{figure}[t]
    \centering
    \includegraphics[width=\linewidth]{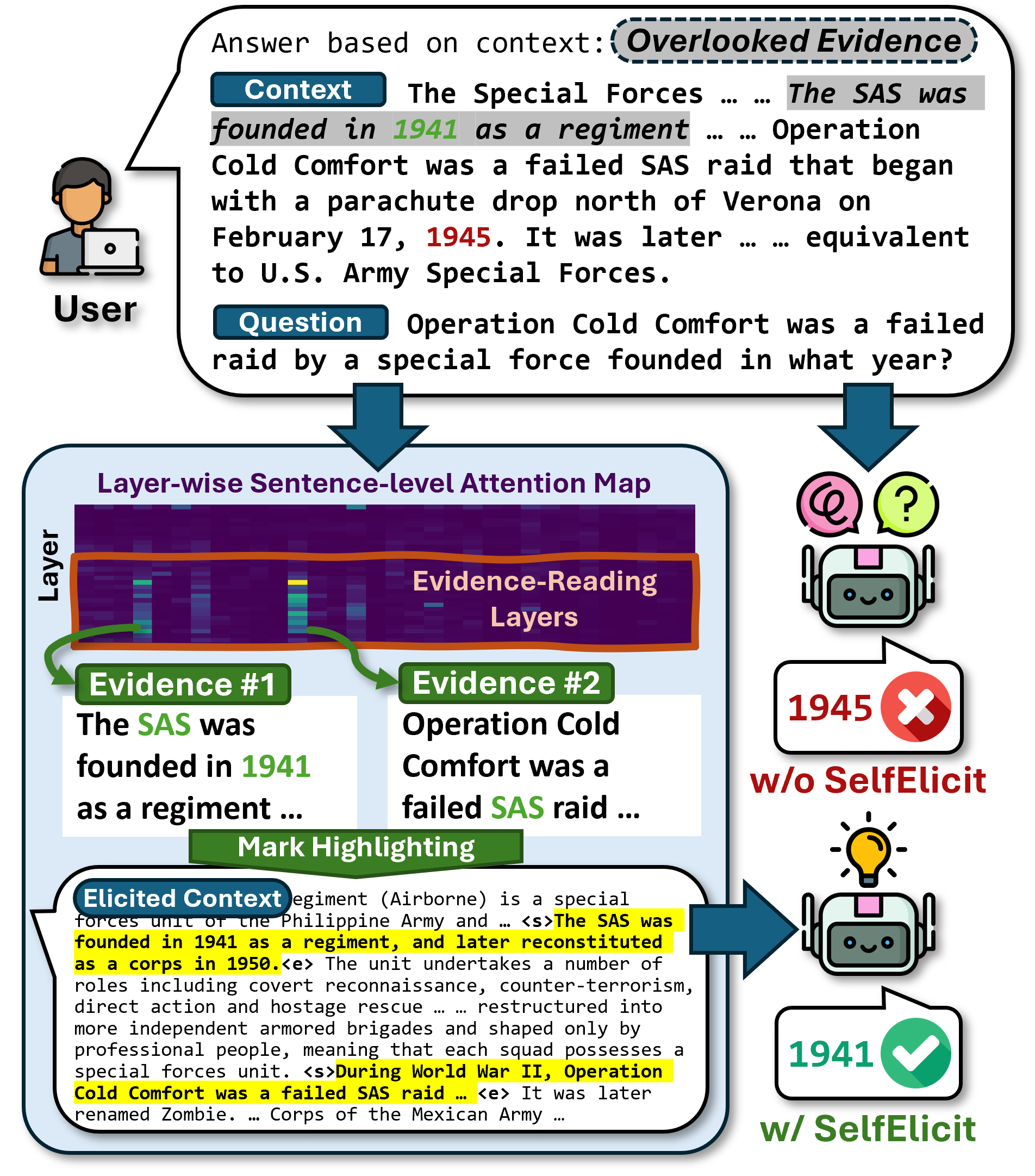}
    \caption{
        \se workflow on a real example with Llama3.1-8B.
        By locating and explicitly highlighting the initially overlooked 2nd-hop evidence (“SAS was founded in 1941 ...") within the context, \se guides the model to arrive at the correct answer “1941”.
    }
    \label{fig:intro}
    \vspace{-10pt}
\end{figure}
However, recent studies find that LMs can fail to properly leverage supporting facts within context, leading to incorrect answers despite available evidence~\cite{shi2024cad,zhao2024cd}.
Such failure largely stems from the noise and irrelevant information in context~\cite{wu2024ragirrelevant, cuconasu2024ragnoise}, which is usually inevitable in practice~\cite{gao2023ragsurvey}.

Recently, improved prompting~\cite{zhou2023prompt} or decoding~\cite{shi2024cad} methods have been proposed to enhance the focus on context. 
However, they treat the whole context as a single entity, overlooking the fact that not all information provided in the context is important.
Meanwhile, another line of work highlights the crucial role of attention mechanisms in identifying key contextual information, leveraging attention patterns for KV cache compression~\cite{li2024snapkv} and document reordering~\cite{peysakhovich2023attentionsort}.
Motivated by these insights, we delve deeper into the capabilities of attention-based mechanisms for fine-grained, inference-time evidence highlighting, enhancing LMs' ability to concentrate on key supporting information within the context.
Our contributions in this work are as follows:
\begin{itemize}
    \item By analyzing the attention scores during response generation, we demonstrate that the LMs have an inherent ability to identify the relevant evidence in the context, regardless of whether they respond correctly or not. This observation holds across various LM families.
    \item Leveraging this inherent ability, we propose an inference time context augmentation approach, \se, that highlights the important evidence in the context and improves LM's ability to provide correct and well-grounded responses. \se is \textbf{efficient} for inference, \textbf{training-free}, and \textbf{robust} to noise and hyper-parameter choice.
    \item In a comprehensive study, we show the significant performance improvement that \se brings across model families and tasks. 
    We also offer extensive studies and analyses on various characteristics of \se, including but not limited to its evidence-eliciting accuracy, effect of different design choices, robustness under noisy/informative contexts, runtime efficiency, etc.
\end{itemize}
Fig.~\ref{fig:intro} illustrates the \se workflow. 
By identifying and highlighting key evidence within the context, \se guides the LM towards important information that it might otherwise overlook, thus leading to the correct response.
\section{Preliminaries}\label{sec:bac}
\paragraph{Problem description.}
Given an LM $\Phi$, question $\vq$, context $\vc$ and QA prompt template $\tau_\text{QA}$, we obtain the generated answer $\vg$ for the question by combining context and question as input: $\vg \gets \Phi(\tau_\text{QA}(\vc, \vq))$.
We use the following template as baseline to test different LMs' base ability in leveraging contextual evidence for QA tasks:
\begin{tcolorbox}[title={\footnotesize A direct prompt template $\tau_\texttt{QA}$ for context-based QA},top=1mm,bottom=1mm]
\scriptsize
Directly answer the question based on the context passage, no explanation is needed. If the context does not contain any evidence, output ``I cannot answer based on the given context."\\
Context: \{\texttt{context}\}\\
Question: \{\texttt{question}\}
\end{tcolorbox}

\paragraph{Notation.}
We only outline the notation and details of Transformer models needed for this work, please see \cite{transformers2017} for more detailed exposition on transformer architecture. Given an input sequence with $n$ tokens, a decoder only transformer LM generates next token by adaptively attending to all previous n tokens. Denote the attention probability vector for attention head $h$ and layer $\ell$ by $\va^{(\ell,h)}\in \R^n$. Then we define
\begin{equation}
\label{eq:att-layer}
\va^{(\ell)} = \frac{1}{H}\sum\limits_{h=1}^{H} \va^{(\ell, h)}
\end{equation}
where $H$ is the total number of heads in a layer and $L$ is the number of layers in the transformer model. Intuitively, $\va^{(\ell)}$ provides a holistic view of layer $\ell$'s attention distribution over the input sequence, summarizing which tokens the model deems most informative at this stage of processing.
\section{\se}\label{sec:met}

\se is an inference-time context augmentation framework to enhance generative LM's capability to rely on relevant information present in the context in order to reduce hallucinations and provide factually correct grounded response. \se requires \textbf{no additional training}, is \textbf{efficient at inference time} and provides \textbf{significant performance improvement} across models and tasks.

\subsection{Self-guided Contextual Evidence Eliciting}
\label{sec:met-elicit}

\begin{figure*}[t]
\vspace{-5pt}
    \centering
    \includegraphics[width=\linewidth]{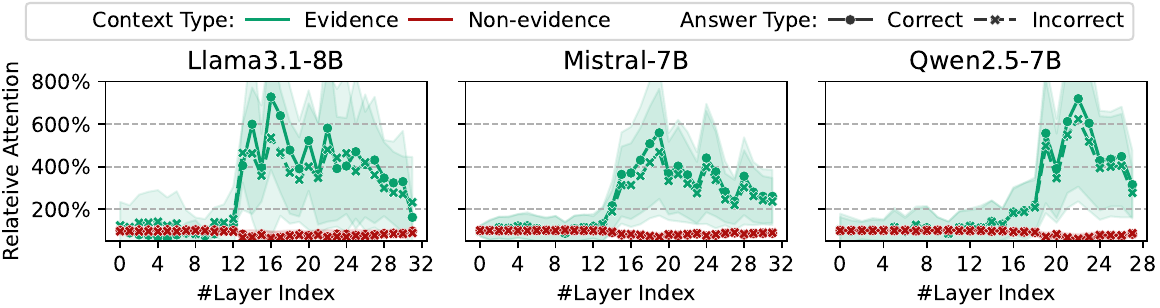}
    \caption[evidence reading layers]{
        Relative attention\footnotemark[3] to the evidence/non-evidence sections (y-axis) across the layers (x-axis) for different LM families on HotpotQA.
        Deeper layers pay much greater attention to crucial evidence (green lines) in the context, even when LM responds incorrectly (dashed lines).
        Best viewed in color.
    }
    \label{fig:layeratt}
\vspace{-5pt}
\end{figure*}

The first step in \se is to leverage the LM's internal representations to identify the relevant evidence sentences in the context. Consider the context-based QA setup described in Sec.~\ref{sec:bac}. Denote the number of tokens in input $\tau(\vc, \vq)$ by $n$ and the number of sentences\footnote{The sentences in a document can be extracted with many natural language processing packages, in this paper we use spaCy~\cite{honnibal2017spacy}.} in the context $\vc$ by $m$. Define $\gS: \{\vs_1, \vs_2, \cdots, \vs_m\}$ where $\vs_i$ denotes $i$-th sentence. Let $(t_{\vs_i}^\text{start}, t_{\vs_i}^\text{end})$ denote the start and end token index of $i$-th sentence. Define sentence-level attention vector $\bva^{(\ell)} \in \R^m$ from token level attention $\va^{(\ell)}$ as follows
\begin{equation}
\label{eq:att-sent}
    \begin{aligned}
    &\bva^{(\ell)} := [\bar{a}_1^{(\ell)}, \bar{a}_2^{(\ell)}, \cdots, \bar{a}_m^{(\ell)}] \in \R^m, \\
    \text{where} \quad & \bar{a}_i^{(\ell)} = \frac{1}{|t_{\vs_i}^\text{end} - t_{\vs_i}^\text{start} + 1|}\sum\limits_{j=t_{\vs_i}^\text{start}}^{t_{\vs_i}^\text{end}} \va^{(\ell)}_j.    
    \end{aligned}
\end{equation}
Intuitively $\bar{a}_i^{(\ell)}$ tells the relative importance of each context sentence $\vs_i$ at layer $\ell$. In Fig.~\ref{fig:layeratt} we plot how $\bar{a}_i^{(\ell)}$ varies across layers for evidence vs non-evidence sentences in the context when generating the first response token. We can clearly see that, regardless of whether the model responds correctly or not, the deeper layers of the LM pay significantly higher attention to the relevant evidence\footnote{In Appendix~\ref{sec:apd-res-attlayer} we explain how we obtain the ground truth for relevant evidence in the context.} in the context. This observation holds across models families and datasets, see Appendix~\ref{sec:apd-res-attlayer}.

We leverage this insight to define the \textbf{sentence evidence scores} $e_i$ for each sentence $\vs_i$ in $\gS$ , which we use at the inference time to predict the evidence containing sentences in the context, as follows: We use a pre-selected subset of layers, denoted by $\gL_\text{ER}$ and referred to as \textbf{evidence-reading} layers, and aggregate $\bar{a}_i^{(\ell)}$ to get $e_i$, i.e., 
\begin{equation}
\label{eq:evd-score}
    \begin{aligned}
    &\ve := [e_1, e_2, \cdots, e_m] \in \R^m, \\
    \text{where} \quad & e_i =  \frac{1}{|\gL_{ER}|} \sum\limits_{\ell \in \gL_{ER}} \bar{a}_i^{(\ell)}
    \end{aligned}
\end{equation}
In Section~\ref{sec:exp-analysis} (RQ4) we study different choices of $\gL_\text{ER}$ for their ability to identify the relevant evidence. We find that using the last 50\% of the layers as $\gL_\text{ER}$ performs consistently well across models and datasets so we use it as default for experiments.

\footnotetext[3]{
Relative attention to the evidence/non-evidence section is defined as the ratio of section-average attention per token (APT) to the context-average APT, e.g., a 600\% evidence relative attention means that LM is paying 6x attention to each token in evidence sentences compared to context average.
}

Next, we introduce an evidence thresholding parameter $\alpha \in [0, 1]$ and use it in combination $e_i$ to predict evidence sentences $\gS_\text{SE}$
\begin{equation}
\label{eq:threshold}
    \gS_\text{SE} = \{\vs_i | \vs_i \in \gS; e_i \geq \alpha \cdot \text{max}(\ve)\},
\end{equation}
A higher value of $\alpha$ will result in fewer sentences being highlighted, and vice versa. It can be seen as a trade-off parameter between precision and recall in evidence elicitation, which can be tuned based on task requirements. \se is generally robust to $\alpha$, see Section~\ref{sec:exp-analysis} (RQ5), with $\alpha=0.5$ yielding significant performance improvement over baseline across models and tasks. Therefore, We use $\alpha=0.5$ as our default setting for experiments. 

We note that similar analyses can be presented at a token level to identify evidence tokens instead of sentences. However, as discussed in Appendix~\ref{sec:apd-res-token}, augmenting the original context by highlighting complete sentences is semantically more meaningful and beneficial than individual tokens.
\subsection{Contextual Evidence Highlighting}
\label{sec:met-highlight}
The second step of \se is to modify the original context to highlight the evidence sentences identified in Sec.~\ref{sec:met-elicit} and to modify the original instructions in Sec.~\ref{sec:bac} to guide the model to use the highlighted evidence. Prioritizing simplicity and efficiency, we introduce a prompt augmentation strategy to achieve this. Specifically for highlighting, in the raw context passage $\vc$ we place text markers "<start\_important>" and "<end\_important>" before and after each sentence in $\gS_\text{SE}$, resulting in the new context $\vc^*$. Furthermore, we update the task instructions to guide the LM's attention towards the highlighted sentences. The response is generated using the mark-highlighted $\vc^*$ and the updated prompt template $\tau_\texttt{SEQA}$.
\begin{tcolorbox}[title={\footnotesize \se Prompt Template $\tau_\texttt{SEQA}$},top=1mm,bottom=1mm]
\scriptsize
\{Original QA Instructions\}
Within the context, <start\_important> and <end\_important> are used to mark the important evidence sentences, read carefully. Do not include the markers in the output.\\
Context: \{\texttt{elicited\_context}\}\\
Question: \{\texttt{question}\}
\end{tcolorbox}
The overall procedure is summarized in Algorithm~\ref{alg:se}.
We note that \se is designed to prioritize efficiency and simplicity, generating only one additional token during inference to identify $\gS_\text{SE}$ (line 1 in Alg.~\ref{alg:se}). 
The highlighting method also outperforms more complex strategies, detailed further in Appendices \ref{sec:apd-highlighting} and \ref{sec:apd-filco}.

\begin{algorithm}[H]
\caption{\se}
\label{alg:se}
\renewcommand{\algorithmicrequire}{\textbf{Input}}
\renewcommand{\algorithmicensure}{\textbf{Output:}}
\begin{algorithmic}[1]
    \REQUIRE{Language model $\Phi$, question $\vq$, context $\vc$, prompt templates $\tau_\text{QA}$ and $\tau_\text{SEQA}$, evidence-reading layers $\gL_\text{ER}$, eliciting threshold $\alpha$.}
    \STATE Generate one token with $\Phi(\tau_\text{QA}(\vc, \vq))$;
    \STATE Compute $\bva^{(\ell)}$ for all $l \in \gL_\text{ER}$; \hfill $\rhd$ \eqref{eq:att-sent}
    \STATE Get evidence score $\ve$ with $\gL_\text{ER}$; \hfill $\rhd$ \eqref{eq:evd-score}
    \STATE Select evidence sentences $\gS_\text{SE}$; \hfill $\rhd$ \eqref{eq:threshold}
    \STATE Derive new context $\vc^*$ by highlighting $\gS_\text{SE}$;
    \STATE Generate answer $\vg_\text{SE} \gets \Phi(\tau_\text{SEQA}(\vc^*, \vq))$
    \ENSURE {The final answer $\vg_\text{SE}$}
\end{algorithmic}
\end{algorithm}
\vspace{-10pt}
\section{Experiments}\label{sec:exp}
We conduct comprehensive experiments across six LMs from different families with varying sizes and on four single- and multi-hop reasoning open-book QA tasks from various domains to investigate:
\vspace{-2mm}
\begin{itemize}[leftmargin=*]
    \setlength{\itemsep}{-3pt}
    \item \textbf{RQ1:} How does \se perform in terms of improving answer quality and accuracy?
    \item \textbf{RQ2:} How do the evidence scores and the context sentences highlighted by \se correlate with relevant evidence?
    \item \textbf{RQ3:} Robustness to noise in the context?
    \item \textbf{RQ4\&5:} How do different choices of evidence-reading layers \& threshold $\alpha$ affect \se?
\end{itemize}
\vspace{-2mm}

\begin{table*}[t]
\centering
\caption{
    Results of applying different evidence-eliciting methods to 6 LMs across 4 context-based QA tasks. 
    We report EM and Token F1 scores (in $\times10^{-2}$) with the gains over direct QA.
    The "average" columns present the average QA performance and the inference time (per sample) across all datasets.
    The best results are \textbf{bolded}.
}
\label{tab:main}
\vspace{-5pt}
\resizebox{\textwidth}{!}{%
\begin{tabular}{ccl|cccccccc|ccc}
\toprule
\multicolumn{2}{c}{\multirow{3}{*}{\textbf{Model}}} & \multicolumn{1}{c|}{\multirow{3}{*}{\textbf{Method}}} & \multicolumn{8}{c|}{\textbf{Dataset}} & \multicolumn{3}{c}{\textbf{Average}} \\ \cline{4-14} 
\multicolumn{2}{c}{} & \multicolumn{1}{c|}{} & \multicolumn{2}{c|}{\textbf{HotpotQA}} & \multicolumn{2}{c|}{\textbf{NewsQA}} & \multicolumn{2}{c|}{\textbf{TQA}} & \multicolumn{2}{c|}{\textbf{NQ}} & \multicolumn{2}{c|}{\textbf{Ranking}} & \textbf{Inference} \\
\multicolumn{2}{c}{} & \multicolumn{1}{c|}{} & \textbf{EM} & \multicolumn{1}{c|}{\textbf{Token F1}} & \textbf{EM} & \multicolumn{1}{c|}{\textbf{Token F1}} & \textbf{EM} & \multicolumn{1}{c|}{\textbf{Token F1}} & \textbf{EM} & \textbf{Token F1} & \textbf{EM} & \multicolumn{1}{c|}{\textbf{Token F1}} & \textbf{Time (ms)} \\ \hline
\multirow{10}{*}{\textbf{\rotatebox{90}{Llama-3.1}}} & \multicolumn{1}{c|}{\multirow{5}{*}{\textbf{\rotatebox{90}{8B}}}} & \textsc{Base} & 58.9 & \multicolumn{1}{c|}{57.7} & 64.3 & \multicolumn{1}{c|}{56.2} & 72.8 & \multicolumn{1}{c|}{66.1} & 59.7 & 61.6 & 4.38 & \multicolumn{1}{c|}{4.75} & 224.1 \\
 & \multicolumn{1}{c|}{} & \textsc{Cot} & 60.4 & \multicolumn{1}{c|}{58.6} & 64.9 & \multicolumn{1}{c|}{55.8} & 74.4 & \multicolumn{1}{c|}{67.4} & 59.6 & 62.1 & 3.75 & \multicolumn{1}{c|}{4.00} & 224.8 \\
 & \multicolumn{1}{c|}{} & \textsc{FullElicit} & 60.7 & \multicolumn{1}{c|}{59.0} & 65.9 & \multicolumn{1}{c|}{56.6} & 72.8 & \multicolumn{1}{c|}{66.3} & 61.1 & 62.5 & 3.12 & \multicolumn{1}{c|}{3.25} & 226.3 \\
 & \multicolumn{1}{c|}{} & \textsc{PromptElicit} & 66.3 & \multicolumn{1}{c|}{66.3} & 62.8 & \multicolumn{1}{c|}{56.7} & 76.0 & \multicolumn{1}{c|}{69.6} & 61.8 & 65.6 & 2.75 & \multicolumn{1}{c|}{2.00} & 1672.0 \\
 & \multicolumn{1}{c|}{} & \cem{\bf\se} & \cem\bs{68.5} & \multicolumn{1}{c|}{\cem\bs{69.5}} & \cem\bs{66.9} & \multicolumn{1}{c|}{\cem\bs{60.8}} & \cem\bs{79.4} & \multicolumn{1}{c|}{\cem\bs{72.7}} & \cem\bs{64.0} & \cem\bs{67.8} & \cem\bs{1.00} & \multicolumn{1}{c|}{\cem\bs{1.00}} & \cem264.1 \\ \cline{2-14} 
 & \multicolumn{1}{c|}{\multirow{5}{*}{\textbf{\rotatebox{90}{70B}}}} & \textsc{Base} & 71.8 & \multicolumn{1}{c|}{74.2} & 66.7 & \multicolumn{1}{c|}{57.4} & 78.0 & \multicolumn{1}{c|}{71.2} & 59.3 & 63.2 & 3.25 & \multicolumn{1}{c|}{3.00} & 1389.8 \\
 & \multicolumn{1}{c|}{} & \textsc{Cot} & 72.4 & \multicolumn{1}{c|}{74.0} & 67.2 & \multicolumn{1}{c|}{56.4} & 77.0 & \multicolumn{1}{c|}{69.9} & 60.3 & 63.1 & 3.12 & \multicolumn{1}{c|}{4.50} & 1394.2 \\
 & \multicolumn{1}{c|}{} & \textsc{FullElicit} & 71.3 & \multicolumn{1}{c|}{73.8} & 66.2 & \multicolumn{1}{c|}{56.8} & 77.5 & \multicolumn{1}{c|}{70.7} & 58.2 & 61.8 & 4.25 & \multicolumn{1}{c|}{4.50} & 1408.1 \\
 & \multicolumn{1}{c|}{} & \textsc{PromptElicit} & 73.4 & \multicolumn{1}{c|}{76.2} & 63.1 & \multicolumn{1}{c|}{58.2} & 77.0 & \multicolumn{1}{c|}{72.3} & 64.0 & 68.5 & 3.38 & \multicolumn{1}{c|}{2.00} & 8124.0 \\
 & \multicolumn{1}{c|}{} & \cem{\bf\se} & \cem\bs{75.9} & \multicolumn{1}{c|}{\cem\bs{79.0}} & \cem\bs{69.2} & \multicolumn{1}{c|}{\cem\bs{62.1}} & \cem\bs{80.0} & \multicolumn{1}{c|}{\cem\bs{74.4}} & \cem\bs{65.4} & \cem\bs{68.7} & \cem\bs{1.00} & \multicolumn{1}{c|}{\cem\bs{1.00}} & \cem1566.9 \\ \hline
\multirow{10}{*}{\textbf{\rotatebox{90}{Mistral}}} & \multicolumn{1}{c|}{\multirow{5}{*}{\textbf{\rotatebox{90}{7B}}}} & \textsc{Base} & 70.4 & \multicolumn{1}{c|}{45.6} & 61.4 & \multicolumn{1}{c|}{32.6} & 81.8 & \multicolumn{1}{c|}{47.9} & 65.7 & 29.9 & 3.75 & \multicolumn{1}{c|}{4.25} & 538.4 \\
 & \multicolumn{1}{c|}{} & \textsc{Cot} & 71.4 & \multicolumn{1}{c|}{44.0} & 60.4 & \multicolumn{1}{c|}{31.4} & 82.2 & \multicolumn{1}{c|}{48.2} & \bs{67.5} & 29.3 & 2.25 & \multicolumn{1}{c|}{4.75} & 560.8 \\
 & \multicolumn{1}{c|}{} & \textsc{FullElicit} & 70.6 & \multicolumn{1}{c|}{47.6} & \bs{62.0} & \multicolumn{1}{c|}{34.0} & 81.3 & \multicolumn{1}{c|}{51.5} & 66.8 & 30.5 & 3.00 & \multicolumn{1}{c|}{3.00} & 541.4 \\
 & \multicolumn{1}{c|}{} & \textsc{PromptElicit} & 71.0 & \multicolumn{1}{c|}{\bs{64.1}} & 57.7 & \multicolumn{1}{c|}{\bs{42.2}} & 81.9 & \multicolumn{1}{c|}{\bs{62.3}} & 65.6 & 42.8 & 4.00 & \multicolumn{1}{c|}{\bs{1.25}} & 1877.8 \\
 & \multicolumn{1}{c|}{} & \cem{\bf\se} & \cem\bs{74.3} & \multicolumn{1}{c|}{\cem61.6} & \cem60.6 & \multicolumn{1}{c|}{\cem41.7} & \cem\bs{83.6} & \multicolumn{1}{c|}{\cem61.3} & \cem66.4 & \cem\bs{43.4} & \cem\bs{2.00} & \multicolumn{1}{c|}{\cem1.75} & \cem431.3 \\ \cline{2-14} 
 & \multicolumn{1}{c|}{\multirow{5}{*}{\textbf{\rotatebox{90}{12B}}}} & \textsc{Base} & 59.2 & \multicolumn{1}{c|}{65.9} & 51.9 & \multicolumn{1}{c|}{51.6} & 72.9 & \multicolumn{1}{c|}{68.7} & 54.8 & 61.6 & 5.00 & \multicolumn{1}{c|}{4.75} & 281.9 \\
 & \multicolumn{1}{c|}{} & \textsc{Cot} & 62.0 & \multicolumn{1}{c|}{69.0} & 53.0 & \multicolumn{1}{c|}{52.6} & 75.6 & \multicolumn{1}{c|}{71.1} & 55.3 & 62.0 & 3.25 & \multicolumn{1}{c|}{3.25} & 284.6 \\
 & \multicolumn{1}{c|}{} & \textsc{FullElicit} & 59.8 & \multicolumn{1}{c|}{65.8} & 53.1 & \multicolumn{1}{c|}{51.9} & 73.7 & \multicolumn{1}{c|}{68.8} & \bs{55.5} & 62.6 & 2.88 & \multicolumn{1}{c|}{4.00} & 283.9 \\
 & \multicolumn{1}{c|}{} & \textsc{PromptElicit} & 62.8 & \multicolumn{1}{c|}{\bs{73.1}} & 52.2 & \multicolumn{1}{c|}{56.6} & 79.9 & \multicolumn{1}{c|}{77.6} & \bs{55.5} & 65.7 & 2.38 & \multicolumn{1}{c|}{1.75} & 1455.0 \\
 & \multicolumn{1}{c|}{} & \cem{\bf\se} & \cem\bs{63.6} & \multicolumn{1}{c|}{\cem72.9} & \cem\bs{54.9} & \multicolumn{1}{c|}{\cem\bs{58.6}} & \cem\bs{82.6} & \multicolumn{1}{c|}{\cem\bs{79.9}} & \cem55.3 & \cem\bs{66.0} & \cem\bs{1.50} & \multicolumn{1}{c|}{\cem\bs{1.25}} & \cem339.1 \\ \hline
\multirow{10}{*}{\textbf{\rotatebox{90}{Qwen2.5}}} & \multicolumn{1}{c|}{\multirow{5}{*}{\textbf{\rotatebox{90}{7B}}}} & \textsc{Base} & 65.2 & \multicolumn{1}{c|}{65.8} & 58.3 & \multicolumn{1}{c|}{45.4} & 77.4 & \multicolumn{1}{c|}{66.1} & 62.2 & 59.9 & 3.75 & \multicolumn{1}{c|}{3.75} & 245.2 \\
 & \multicolumn{1}{c|}{} & \textsc{Cot} & \bs{70.7} & \multicolumn{1}{c|}{37.9} & 59.2 & \multicolumn{1}{c|}{31.9} & \bs{78.6} & \multicolumn{1}{c|}{41.6} & 63.8 & 32.3 & 2.00 & \multicolumn{1}{c|}{5.00} & 421.5 \\
 & \multicolumn{1}{c|}{} & \textsc{FullElicit} & 65.5 & \multicolumn{1}{c|}{65.7} & 57.5 & \multicolumn{1}{c|}{48.2} & 77.1 & \multicolumn{1}{c|}{67.3} & 64.4 & 60.6 & 3.50 & \multicolumn{1}{c|}{2.75} & 249.6 \\
 & \multicolumn{1}{c|}{} & \textsc{PromptElicit} & 64.7 & \multicolumn{1}{c|}{67.7} & 54.9 & \multicolumn{1}{c|}{46.3} & 75.4 & \multicolumn{1}{c|}{66.8} & 64.5 & \bs{65.0} & 4.25 & \multicolumn{1}{c|}{2.25} & 1165.1 \\
 & \multicolumn{1}{c|}{} & \cem{\bf\se} & \cem69.1 & \multicolumn{1}{c|}{\cem\bs{71.4}} & \cem\bs{59.6} & \multicolumn{1}{c|}{\cem\bs{50.8}} & \cem78.1 & \multicolumn{1}{c|}{\cem\bs{67.8}} & \cem\bs{65.0} & \cem64.7 & \cem\bs{1.50} & \multicolumn{1}{c|}{\cem\bs{1.25}} & \cem289.4 \\ \cline{2-14} 
 & \multicolumn{1}{c|}{\multirow{5}{*}{\textbf{\rotatebox{90}{32B}}}} & \textsc{Base} & 71.8 & \multicolumn{1}{c|}{68.4} & 60.0 & \multicolumn{1}{c|}{44.7} & 79.0 & \multicolumn{1}{c|}{69.3} & 62.7 & 59.3 & 3.12 & \multicolumn{1}{c|}{3.25} & 928.2 \\
 & \multicolumn{1}{c|}{} & \textsc{Cot} & 71.3 & \multicolumn{1}{c|}{67.1} & 60.0 & \multicolumn{1}{c|}{43.5} & 79.5 & \multicolumn{1}{c|}{66.8} & 59.5 & 55.3 & 3.62 & \multicolumn{1}{c|}{5.00} & 998.6 \\
 & \multicolumn{1}{c|}{} & \textsc{FullElicit} & 71.3 & \multicolumn{1}{c|}{68.2} & 61.6 & \multicolumn{1}{c|}{45.7} & 78.5 & \multicolumn{1}{c|}{68.8} & 63.2 & 58.1 & 3.25 & \multicolumn{1}{c|}{3.75} & 936.3 \\
 & \multicolumn{1}{c|}{} & \textsc{PromptElicit} & 71.3 & \multicolumn{1}{c|}{74.5} & 59.0 & \multicolumn{1}{c|}{51.5} & 78.0 & \multicolumn{1}{c|}{69.9} & 64.8 & 68.1 & 4.00 & \multicolumn{1}{c|}{2.00} & 5109.8 \\
 & \multicolumn{1}{c|}{} & \cem{\bf\se} & \cem\bs{73.3} & \multicolumn{1}{c|}{\cem\bs{75.0}} & \cem\bs{65.6} & \multicolumn{1}{c|}{\cem\bs{57.3}} & \cem\bs{82.1} & \multicolumn{1}{c|}{\cem\bs{74.8}} & \cem\bs{66.8} & \cem\bs{69.8} & \cem\bs{1.00} & \multicolumn{1}{c|}{\cem\bs{1.00}} & \cem980.7 \\ \bottomrule
\end{tabular}%
}
\end{table*}

\begin{table*}[t]
\scriptsize
\centering
\caption{
Examples of how \se helps LM, with \textev{blue} text highlighting the evidence selected by \se. Due to space limitation we only display a small portion of the full context, please see Appendix~\ref{sec:apd-res-exa} for full context. 
}
\label{tab:example}
\vspace{-5pt}
\resizebox{\textwidth}{!}{%
\begin{tabular}{p{0.01\textwidth} p{0.75\textwidth} p{0.22\textwidth}}
\toprule
{\bf\small} & {\bf\small (Partial) Context Passage} & {\bf\small Question \& Answers}\\
\midrule
\multirow{7}{*}{\rotatebox{90}{\bf True or False}}
& 
\uline{\textbf{(Displaying 123 of 794 Words)}}
\textev{Tiger Please is an Indie / Alternative five-piece band from Cardiff, Wales. The band formed in August 2008.} The band's influences are U2, Sigur Rós, Kings of Leon, John Mayer and Counting Crows. They signed with Walnut Tree Records in 2009 and released their debut EP "They Don't Change Under Moonlight". "Kerrang!" magazine, "Rock Sound" magazine, and "Classic Rock" magazine praised the EP and featured the band on the "Rock Sound" and "Classic Rock" cover-mount albums. ... ... \textev{Black Rebel Motorcycle Club (often abbreviated as BRMC) is an American rock band from San Francisco, California.} The group consists of Peter Hayes (vocal, guitar, harmonica), Robert Levon Been (vocal, bass, guitar), and Leah Shapiro (drums). Former drummer Nick Jago left the band in 2008 ... ...
& 
{\scriptsize
    \textbf{Question:} Are the bands Tiger Please and Black Rebel Motorcycle Club from the same country?\vspace{1mm}
    
    \textbf{True Answer:} No. \vspace{1mm}
    
    \uline{\textbf{\texttt{Base:}}}
    Yes. \xmark \vspace{1mm}
    
    \uline{\textbf{\texttt{+\se:}}}
    No. \cmark \vspace{1mm}
}
\\ \midrule
\multirow{8}{*}{\rotatebox{90}{\bf Comparison}}
& 
\uline{\textbf{(Displaying 129 of 227 Words)}}
\textev{Home Monthly was a monthly women's magazine published in Pittsburgh, Pennsylvania in the late 19th century.} "The Strategy of the Were-Wolf Dog" is a short story by Willa Cather. \textev{It was first published in "Home Monthly" in December 1896.} The Count of Crow's Nest is a short story by Willa Cather. It was first published in "Home Monthly" in October 1896. \textev{Mirabella was a women's magazine published from June 1989 to April 2000.} It was created by and named for Grace Mirabella, a former "Vogue" editor in chief, in partnership with Rupert Murdoch. "Nanette: An Aside" is a short story by Willa Cather. It was first published in "Courier" on 31 July 1897 and one month later in "Home Monthly". "The Prodigies" is a short story by Willa Cathe ... ...
& 
{\scriptsize
    \textbf{Question:} Which women's magazine was published first, Mirabella or Home Monthly?\vspace{1mm}
    
    \textbf{True Answer:} Home Monthly. \vspace{1mm}
    
    \uline{\textbf{\texttt{Base:}}}
    Mirabella. \xmark \vspace{1mm}
    
    \uline{\textbf{\texttt{+\se:}}}
    Home Monthly. \cmark \vspace{1mm}
}
\\ \midrule
\multirow{8}{*}{\rotatebox{90}{\bf Fact Retrieval}}
& 
\uline{\textbf{(Displaying 153 of 1014 Words)}}
... ... Lars Lunde (born 21 March 1964) is a Danish former professional football player, who played in the striker position. Lunde got his breakthrough with Brøndby IF in 1983, and he made his debut for the Denmark national football team in October 1983. He was sold to Young Boys Bern in Switzerland, before moving to German club Bayern Munich in 1986. \textev{He was a part of the Bayern team which won the German Bundesliga championship in 1987, and he came on as a late substitute when Bayern lost the 1987 European Cup Final to FC Porto.} He played the last of his three matches for the Danish national team in April 1987, before leaving Bayern during the 1987–88 season. He went on to play for a number of smaller clubs, ending his career with FC Baden in Switzerland. ... ...
& 
{\scriptsize
    \textbf{Question:} Which team did Lars Lunde play for when defeated for the 1987 European Cup Final?\vspace{1mm}
    
    \textbf{True Answer:} Bayern Munich \vspace{1mm}
    
    \uline{\textbf{\texttt{BASE:}}}
    FC Porto. \xmark \vspace{1mm}
    
    \uline{\textbf{\texttt{+\se:}}}
    Bayern Munich. \cmark \vspace{1mm}
}
\\\midrule
\multirow{8}{*}{\rotatebox{90}{\bf Multi-hop Reasoning}}
& 
\uline{\textbf{(Displaying 146 of 532 Words)}}
... ... This sent Olympic down to play in the Premier League in 2007. Adelaide City won the title with games to spare after being runaway leaders, finishing the season unbeaten. \textev{Norwood is a suburb of Adelaide, about 4 km east of the Adelaide city centre.} The suburb is in the City of Norwood Payneham \& St Peters, the oldest South Australian local government municipality, with a city population over 34,000. Whyalla railway station was the terminus station of the Whyalla line serving the South Australian city of Whyalla. \textev{Walter Frank Giffen (20 September 1861 in Norwood – 28 June 1949 in Adelaide) was an Australian cricketer who played in 3 Tests between 1887 and 1892.} He was the brother of the great all-rounder George Giffen. The City of Burnside is a local government area with an estimated population of 44,300 people in the South Australian city of Adelaide. ... ... 
& 
{\scriptsize
    \textbf{Question:} Walter Giffen is from a suburb of which South Australian city?\vspace{1mm}
    
    \textbf{True Answer:} Adelaide\vspace{1mm}
    
    \uline{\textbf{\texttt{Base:}}}
    Norwood. \xmark \vspace{1mm}
    
    \uline{\textbf{\texttt{+\se:}}}
    Adelaide. \cmark \vspace{1mm}
}
\\
\bottomrule
\end{tabular}
}
\vspace{-5pt}
\end{table*}

\subsection{Experimental Setup}
\vspace{-0.5em}

\paragraph{Datasets and Metrics.}
We test 4 datasets: HotpotQA~\cite{yang2018hotpotqa} and the MRQA version \cite{fisch2019mrqa} of NewsQA~\cite{trischler2017newsqa}, TriviaQA (TQA)~\cite{joshi2017triviaqa}, and Natural Questions (NQ)~\cite{kwiatkowski2019natural}.
These datasets feature context passages from diverse sources (e.g., web/Wikipedia/news reports), requiring the model to reason over a single or multiple pieces of evidence within the context. 
This provides a comprehensive test of \se in real-world applications.
For all datasets, we use the official validation split on HuggingFace for testing.
We apply greedy decoding to get deterministic answers. 
Exact Match (EM) and Token-level F1 scores are used for QA performance evaluation.

\paragraph{Models and Baselines.}
We test six open-source instruction fine-tuned models: Llama-3.1 (8B, 70B)~\cite{dubey2024llama}, Mistral (7B, 12B)~\cite{jiang2023mistral}, and Qwen2.5 (7B, 32B)~\cite{yang2024qwen}.
We compare \se to the following prompting/evidence-eliciting approaches:
\begin{itemize}
    \setlength{\itemsep}{0pt}
    \item \co~\cite{wei2022chain}: Chain-of-thought prompting encourages the model to reason through intermediate steps before reaching a final answer. 
    While CoT promotes step-by-step reasoning based on the evidence in the context, it does not explicitly highlight important information within the context. 
    Therefore, we use it as a natural baseline for validating the advantage of explicit evidence elicitation in \se.
    We implement it by adding the \co prompt "Think step by step to provide the answer." at the end of the instruction.
    \item \fe: A naive approach that highlights the entire context as important. Comparing with it shows the necessity of fine-grained, sentence-level evidence elicitation.
    \item \pe: This method leverages the LM itself for generative evidence extraction. 
    It involves two steps: first, the LM is prompted to select the most relevant evidence sentences from the context passage that can help answer the question. 
    Then, we highlight the extracted evidence and get new context for QA following Section~\ref{sec:met-highlight}. 
    Note that this method involves generative evidence extraction in iterative prompting, which requires the model to generate a large amount of additional tokens.
    This serves as a strong baseline for evaluating whether the quality of evidence selected by \se can match that of evidence extracted through generative approach. We use the following prompt for generative evidence extraction step:
\end{itemize}
\vspace{-5pt}
\begin{tcolorbox}[title={\footnotesize \pe Prompt Template $\tau_\texttt{PE}$},top=1mm,bottom=1mm]
\scriptsize
Please find the supporting evidence sentences from the context for the question, then copy-paste the original text to output. Template for output: `- [sentence1] - [sentence2] ...'\\
Context: \{\texttt{context}\}\\
Question: \{\texttt{question}\}
\end{tcolorbox}
Since our focus is on analyzing and improving the inherent ability of the generative LM to provide factually grounded responses based on relevant information provided in the context, we do not compare to any dataset specific and task specific fine-tuning methods. For a fair comparison, we do not tune the hyper-parameters for \se. They are fixed as $\gL_\text{ER}$ being the last 50\% layers and $\alpha=0.5$.

\subsection{Main Results}\label{sec:exp-main}
\vspace{-0.5em}

\paragraph{RQ1: \se consistently improves grounded factuality.}
Table~\ref{tab:main} compares \se to the other methods. We observe that
\begin{itemize}
    \item As expected \fe does not provide meaningful and consistent improvement since it doesn't elicit fine-grained evidence. 
    \item \co also does not provide a meaningful and consistent gain which highlights the fact that asking the model to reason carefully does not improve its ability to leverage the relevant evidence while generating the response. \co results also show variation across LM families where we don't observe meaningful gains for Llama and Mistral models. When combined with Qwen models, \co generates significantly longer answers with intermediate steps. 
    While this results in a slight increase in EM scores in a few cases, it also leads to a large drop (up to 27.9) in F1 scores due to the inclusion of redundant information, as well as longer inference times (e.g., 171.9\% more inference time for Qwen2.5-7B).
    \item \se significantly and consistently improves the performance across all datasets and models of different sizes (5.0\%-11.7\% gain over baseline). Even when compared to computationally expensive (average inference time increase of 878\%/939\% for Llama3.1-8B/70B) iterative prompting approach \pe, \se outperforms for 40 out of 48 model-task metric pairs while incurring a fraction of the computational cost increment (only $\sim$3-5\% increase when compared to \pe).

\end{itemize}
Appendix~\ref{sec:apd-res-eff} provides a more detailed discussion of computational efficiency of the approaches.

\paragraph{RQ2: \se highlights relevant evidence.}
We demonstrate both qualitatively and quantitatively how the context highlighted by \se is the relevant evidence for the task which leads to the performance gains reported in Table~\ref{tab:main}.

For qualitative illustration, in Table~\ref{tab:example} we show various types of examples (e.g., true/false, comparison, fact retrieval, multi-hop reasoning) from the HotpotQA dataset. \se identifies the most relevant supporting facts across different QA tasks. For instance, in the 2$^\text{nd}$ "comparison" example that asks which magazine was published first, \se highlights the information stating the publication dates of the magazines and related articles. In the 4$^\text{th}$ "multi-hop reasoning" example, it emphasizes the overlooked 2$^\text{nd}$-hop evidence “Norwood is a suburb of Adelaide”, which helps the model arrive at the correct answer “Adelaide”.

\begin{figure*}[t]
  \subfigure[
    Performance of base and \se-augmented LM with/without distracting context information, evaluated by Exact Match and Token F1 score.
  ]{
  \label{fig:noise-qa}
    \centering
    \includegraphics[width=0.58\linewidth]{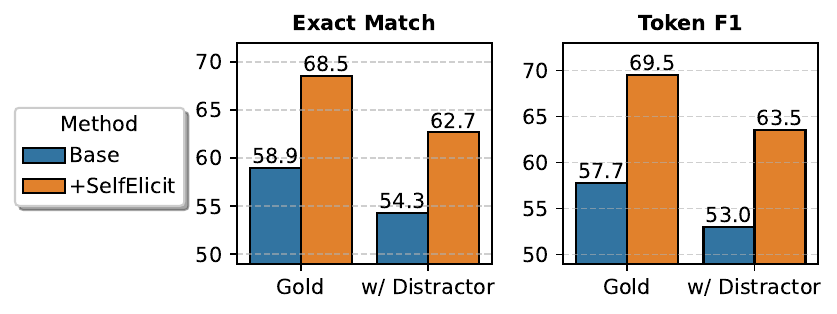}
  }
\hfill
  \subfigure[
    Context elicitation ratio of \se with/without distracting context information.
  ]{
  \label{fig:noise-elicit-ratio}
    \centering
    \includegraphics[width=0.35\linewidth]{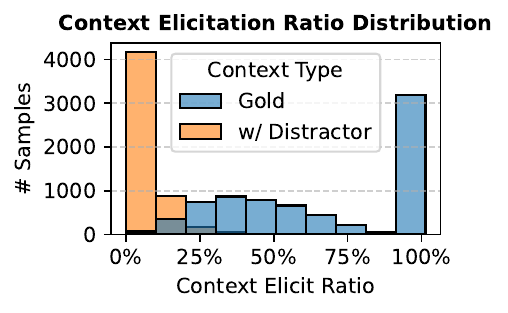}
  }
  \vspace*{-5pt}
  \caption{
    \se demonstrates robust advantage even in the presence of noisy context (Fig.~\ref{fig:noise-qa}).
    When the context passages contain more distracting information, \se tends to select a significantly smaller portion of text as evidence (Fig.~\ref{fig:noise-elicit-ratio}) to prevent the LM from being distracted by irrelevant contexts.
  }
  \label{fig:noise}
\end{figure*}

\begin{table}[h]
\footnotesize
\centering
\caption{\se accurately identifies contextual evidence sentences across different datasets and models.}
\label{tab:elicit}
\vspace{-5pt}
\resizebox{0.8\columnwidth}{!}{%
\begin{tabular}{@{}cccc@{}}
\toprule
\multicolumn{1}{c|}{\multirow{2}{*}{\textbf{Dataset}}} & \multicolumn{3}{c}{\textbf{Model}} \\
\multicolumn{1}{c|}{} & \textbf{Llama3.1} & \textbf{Mistral} & \textbf{Qwen2.5} \\ \midrule
\multicolumn{4}{c}{\textbf{Metric: Sentence-level AUROC}} \\ \midrule
\multicolumn{1}{c|}{HotpotQA} & 91.24 & 85.35 & 88.21 \\
\multicolumn{1}{c|}{NewsQA} & 92.68 & 88.68 & 91.54 \\
\multicolumn{1}{c|}{TQA} & 73.27 & 68.89 & 70.59 \\
\multicolumn{1}{c|}{NQ} & 90.87 & 85.51 & 87.43 \\ \midrule
\multicolumn{4}{c}{\textbf{Metric: Sentence-level NDCG}} \\ \midrule
\multicolumn{1}{c|}{HotpotQA} & 91.36 & 82.45 & 87.05 \\
\multicolumn{1}{c|}{NewsQA} & 82.79 & 70.65 & 82.37 \\
\multicolumn{1}{c|}{TQA} & 66.41 & 63.32 & 67.19 \\
\multicolumn{1}{c|}{NQ} & 91.45 & 86.45 & 87.65 \\ \bottomrule
\end{tabular}%
}
\vspace{-10pt}
\end{table}

For quantitative evaluation, we assess the accuracy of evidence elicitation by checking whether \se assigns higher evidence scores to \textit{ground-truth} evidence sentences. 
Specifically, for HotpotQA, we use the "supporting\_facts" annotations to derive ground-truth evidence labels, while for other datasets, a sentence is treated as ground truth evidence if it contains at least one of the correct answers.
Since \se computes \textit{continuous} evidence scores for each sentence (see \eqref{eq:evd-score}), we utilize AUROC and NDCG@all to assess the score for accurately classifying/ranking contextual evidence.
Table~\ref{tab:elicit} shows that SE can accurately locate evidence sentences across models and datasets, with 80-95 AUROC or NDCG scores in most cases. We note that results on the TQA dataset are relatively lower because TQA often has multiple candidate correct answers and thus a larger set of evidence. During inference, the model typically focuses on evidence for only one of the answers, which reduces the evidence score of other (candidate) evidence sentences. However, this does not hinder \se's ability to help LM respond better, as demonstrated by the results in Table~\ref{tab:main}.

\subsection{Additional Analysis}\label{sec:exp-analysis}
\vspace{-0.5em}

\paragraph{RQ3: \se is effective in presence of context noise.}
To explore the impact of real world noise on \se we study it's performance for the "distractor" variant of HotpotQA dataset containing additional distracting information, retrieved from Wikipedia~\cite{yang2018hotpotqa}, in the context. The distractor setting increases the average length of the context by 1443\% by introducing irrelevant information. Figure~\ref{fig:noise-qa} shows that \se maintains the advantage over the baseline even in the presence of substantial context noise. 

However, similar to the baseline, the performance of \se for "distractor" variant worsens meaningfully when compared to \se performance for the gold context setup, also shown in Figure~\ref{fig:noise-qa}. To investigate this we do a deep dive into the evidence elicitation capability of \se in the presence of substantial noise. Figure~\ref{fig:noise-elicit-ratio} contrasts the elicit ratio (i.e., the proportion of elicited evidence within the original context) under gold and distractor settings. 
With gold context, \se tends to select a larger proportion as evidence since most contextual information provided is relevant. By contrast, in the distractor setting where most of the context is irrelevant, the proportion of evidence selected decreases substantially (typically <10\%) for the same $\alpha=0.5$.
This shows that even with a static $\alpha$, \se naturally exhibits adaptiveness by selecting a lower evidence ratio in noisy contexts, focusing only on the relevant evidence without distraction from noise.

\paragraph{RQ4: Deeper Layers are better for $\gL_\text{ER}$.}
We already saw in Fig.~\ref{fig:layeratt} that the deeper layers of all the model families exhibit higher attention scores and a clear ability to distinguish relevant evidence within the context.  In Table~\ref{tab:layer-choice} we verify this further empirically by comparing the evidence elicit accuracy and QA performance of seven different choices of $\gL_\text{ER}$ for Llama3.1-8B evaluated on HotpotQA with $\alpha=0.5$. We see that, consistent with the observation in Fig.~\ref{fig:layeratt}, choosing last 50\% of the layers as $\gL_\text{ER}$ leads to (close to) best metrics in terms of both evidence elicitation and task performance. Based on the qualitative observation in Fig.~\ref{fig:layeratt}, this choice also looks robust across model families which is then further verified by the results in Table~\ref{tab:main} where we see a consistent significant improvement across all model families and tasks for this choice of $\gL_\text{ER}$. While there may be other choices of $\gL_\text{ER}$ which lead to better performance gains for a specific model-dataset pair, finding them requires model and task specific hyper-parameter tuning unlike universally applicable default setting.

\begin{table}[h]
\centering
\caption{Effect of different evidence-reading (ER) layer choices on elicit accuracy and QA performance.}
\label{tab:layer-choice}
\vspace{-5pt}
\resizebox{\columnwidth}{!}{%
\begin{tabular}{@{}c|cc|cc@{}}
\toprule
\multirow{2}{*}{\textbf{ER Layer Span}} & \multicolumn{2}{c|}{\textbf{Elicit Accuracy}} & \multicolumn{2}{c}{\textbf{QA Performance}} \\
 & \textbf{AUROC} & \textbf{NDCG} & \textbf{EM} & \textbf{Token F1} \\ \midrule
0\%-100\% & 89.02 & 75.50 & 62.57 & 63.33 \\ \midrule
0\%-50\% & 70.38 & 44.99 & 62.14 & 62.65 \\
50\%-100\% & \sbs{91.55} & \bs{80.11} & \bs{64.86} & \bs{65.23} \\ \midrule
0\%-25\% & 59.01 & 37.55 & 61.86 & 61.83 \\
25\%-50\% & 74.82 & 48.80 & 62.57 & 62.73 \\
50\%-75\% & \bs{91.66} & \sbs{79.96} & \sbs{63.57} & \sbs{64.14} \\
75\%-100\% & 91.02 & 78.72 & 63.43 & 64.10 \\ \bottomrule
\end{tabular}%
}
\vspace{-10pt}
\end{table}

\begin{figure*}[t]
    \centering
    \includegraphics[width=\linewidth]{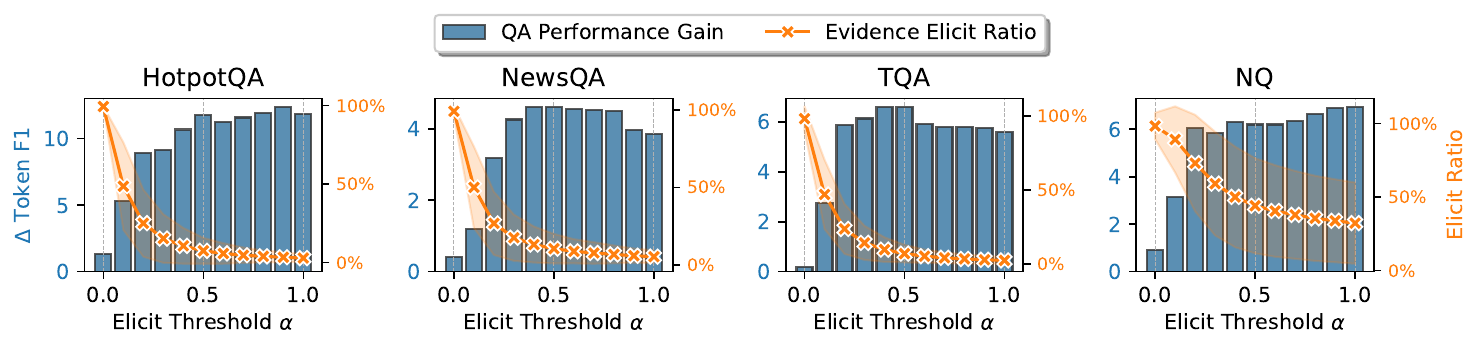}
    \vspace{-15pt}
    \caption{
        Impact of elicit threshold $\alpha$ (x-axis) on the QA performance gain (blue bars, left y-axis) and evidence elicit ratio (orange lines, right y-axis) of \se on four QA tasks.
        Best viewed in color.
    }
    \label{fig:threshold}
\end{figure*}

\paragraph{RQ5: Balancing elicitation precision and comprehensiveness in choice of $\alpha$.}
The choice of $\alpha$ in \se acts as a proxy for trade-off between evidence elicitation precision vs coverage. For example, for $\alpha=1$ we only select the sentence with highest $e_i$ whereas for $\alpha=0$ we have the whole context selected, analogue to \fe. In Figure~\ref{fig:threshold} we show the impact of $\alpha$ on the Token F1 score and evidence elicitation ratio for Llama3.1-8B on four QA datasets. We observe
\begin{itemize}
    \setlength{\itemsep}{-3pt}
    \item For all datasets the performance quickly rises and the evidence elicitation ratio quickly drops with increasing $\alpha$ for smaller values of $\alpha$ indicating clearly that evidence elicitation upto a threshold helps all datasets. 
    \item Beyond $\alpha=0.5$ there is relatively minor variation in the performance with change in $\alpha$, illustrating robustness of \se to choice of $\alpha \in [0.5, 1]$ across datasets. The minor variation is dependent on the nature of the dataset. For datasets that require multi-hop reasoning, relying on multiple pieces of relevant evidence, we achieve optimal performance for $\alpha$ closer to $0.5$ whereas for NQ requiring simpler reasoning we achieve the best performance for $\alpha=1$. 
\end{itemize}
To strike a balanced trade-off between evidence elicitation precision and coverage, we choose $\alpha=0.5$ as default for experiments which leads to consistent gains across models and datasets (Table~\ref{tab:main}).
\section{Related Works}\label{sec:rel}
\vspace{-5pt}
\paragraph{Context-based question answering.}
Furnishing LMs with relevant context is an effective way of providing up-to-date external and/or private knowledge~\cite{ji2023survey,selfrag} to help mitigate hallucination and improve response accuracy~\cite{hallucination_survey}. 
Retrieval-augmented Generation (RAG) is a widely adopted paradigm for this purpose~\cite{gao2023ragsurvey,fan2024ragsurvey}.
Despite its popularity, recent studies have pointed out that context retrieved from external sources often contains noise and irrelevant information, leading to confusion for LM~\cite{cuconasu2024ragnoise,wu2024ragirrelevant,yu-etal-2024-popalm}.
Motivated by this, we explore how to leverage contextual information more effectively at a finer granularity by highlighting critical information within the context.
To the best of our knowledge, we are the first to investigate automated contextual evidence highlighting based on LM internal representations.

\paragraph{Factuality and internal representation.}
Recent studies have explored ways to understand QA factuality by analyzing the internal representations to identify important attention layers~\cite{yuksekgonul2023attention,chen2024sharpness} or heads~\cite{halawi2024overthinking} that are crucial for generation correctness or hallucination.
However, they primarily focus on how LMs utilize their parametric knowledge in controlled generation~\cite{yuksekgonul2023attention,halawi2024overthinking} or closed-book QA~\cite{chen2024sharpness}, with little discussion on context-based QA that rely on using external non-parametric knowledge.
For context-based QA, \citet{peysakhovich2023attentionsort} propose combating positional bias by iteratively ranking retrieved context documents based on average attention scores.
\citet{wang2023filco} employs a fine-tuned Llama-2 evidence extractor to filter context information. 
Nevertheless, these works either employ document-level average attention over all layers and heads or necessitate training an auxiliary model for generative filtering.
In contrast, our work delves into a detailed analysis of the behavior of each attention layer at the sentence level, enabling fine-grained and efficient evidence highlighting at inference time.

\section{Conclusion}\label{sec:con}
\vspace{-5pt}
This paper proposes a novel method \se that boosts LM's ability to utilize context with automatic contextual evidence highlighting.
It harnesses the inherent ability of specific attention layers within the LM to differentiate evidence within the context, enabling test-time auto-eliciting that is general, efficient and requires no additional training.
Comprehensive experiments on context based QA validate the effectiveness of \se, showing a performance improvement of 5.0\% to 11.7\% across multiple datasets and LMs. 
Compared to more costly generative evidence extraction, our method achieves better performance with far less additional computational overhead.

\section*{Limitations}
We evaluated the effectiveness of \se across several open-source LMs. However, we can not assess \se on proprietary LMs due to the need to access attention scores. 
Prioritizing efficiency and simplicity, we used a static eliciting threshold in the paper. 
We discussed the effect of different threshold $\alpha$ choices in Figure~\ref{fig:threshold}.
Our method is generally robust to different threshold $\alpha$, with a broad range of values yielding consistently strong improvements, but one can still fine-tune it on specific tasks for better performance.
Moreover, an interesting future direction would be to automatically select the optimal elicit thresholds based on the model and input characteristics. 
We provide a more detailed case analysis and discussion on the potential benefit of dynamic threshold adjustment in Appendix \ref{sec:apd-failcase}.

\section*{Acknowledgments}
We thank Danai Koutra for her insights and discussions throughout the project.
This work is partially supported by NSF (2134079
) and
AFOSR (FA9550-24-1-0002), %MURI network alignment
The content of the information in this document does not necessarily reflect the position or the policy of the Government, and no official endorsement should be inferred.  The U.S. Government is authorized to reproduce and distribute reprints for Government purposes notwithstanding any copyright notation here on. 

\normalem
\bibliography{custom}

\appendix

\section{Reproducibility Details}
\label{sec:app-rep}

\paragraph{QA pipeline details.}
We use PyTorch~\cite{paszke2019pytorch} and HuggingFace Transformers~\cite{wolf2020transformers} to implement all involved language models (LMs).
Specifically, the LMs used in the paper are the instruct fine-tuned version of Llama3.1 8B\footnote{\url{https://huggingface.co/meta-llama/Llama-3.1-8B-Instruct}} and 70B\footnote{\url{https://huggingface.co/meta-llama/Llama-3.1-70B-Instruct}}, Mistral 7B (instruct-v0.3)\footnote{\url{https://huggingface.co/mistralai/Mistral-7B-Instruct-v0.3}} and 12B (Nemo)\footnote{\url{https://huggingface.co/mistralai/Mistral-Nemo-Instruct-2407}}, as well as Qwen2.5 7B\footnote{\url{https://huggingface.co/Qwen/Qwen2.5-7B-Instruct}} and 32B\footnote{\url{https://huggingface.co/Qwen/Qwen2.5-32B-Instruct}}.
We use the official \texttt{apply\_chat\_template} function to get the input tokens from text input ($\tau(\vc, \vq)$) and let the model generate the answer based on it.
We use greedy decoding with a temperature of 0 to get deterministic answers.

\paragraph{Dataset details.}
For all the datasets used, we test the LM on the official public validation split, with 7,405/4,212/7,785/12,836 context-question pairs for HotopotQA/NewsQA/TQA/NQ.
The context passages in those datasets are retrieved from diverse sources including Wikipedia~\cite{yang2018hotpotqa,kwiatkowski2019natural}, news articles~\cite{trischler2017newsqa}, and web search queries~\cite{joshi2017triviaqa}. 
The detailed statistics and descriptions are reported in Table~\ref{tab:datasets}. We use Llama-3.1 tokenizer to compute the number of tokens in the context passage.

\begin{table}[H]
\vspace{-5pt}
\caption{Dataset description and statistics.}
\label{tab:datasets}
\vspace{-10pt}
\resizebox{\columnwidth}{!}{%
\begin{tabular}{@{}c|cccc@{}}
\toprule
\multirow{2}{*}{\textbf{Dataset}} & \multicolumn{2}{c}{\textbf{\#ContextTokens}} & \multirow{2}{*}{\textbf{\#Samples}} & \multirow{2}{*}{\textbf{Source}} \\
 & \textbf{Avg.} & \textbf{Max} &  &  \\ \midrule
HotpotQA & 1251.6 & 3346 & 7405 & Wikipedia \\
NewsQA & 625.24 & 940 & 4212 & CNN News Article \\
TQA & 892.89 & 1974 & 7785 & Bing Search Query \\
NQ & 249.38 & 2679 & 12836 & Wikipedia \\ \bottomrule
\end{tabular}%
}
\vspace{-10pt}
\end{table}

\begin{table*}[t]
\centering
\caption{Detailed runtime and efficiency results. we report the average inference time and number of generated tokens for each sample (i.e., context-question pair) for each dataset.}
\label{tab:efficiency}
\vspace{-10pt}
\resizebox{\textwidth}{!}{%
\begin{tabular}{@{}ccllllllllllc@{}}
\toprule
\multicolumn{2}{c}{\multirow{3}{*}{\textbf{Model}}} & \multicolumn{1}{c}{\multirow{3}{*}{\textbf{Method}}} & \multicolumn{8}{c}{\textbf{Dataset}} & \multicolumn{2}{c}{\textbf{Average}} \\ \cmidrule(l){4-13} 
\multicolumn{2}{c}{} & \multicolumn{1}{c}{} & \multicolumn{2}{c|}{\textbf{HotpotQA}} & \multicolumn{2}{c|}{\textbf{NewsQA}} & \multicolumn{2}{c|}{\textbf{TQA}} & \multicolumn{2}{c}{\textbf{NQ}} & \multicolumn{1}{c}{\textbf{Inference}} & \textbf{\#Generated} \\
\multicolumn{2}{c}{} & \multicolumn{1}{c}{} & \multicolumn{1}{c}{\textbf{Time (ms)}} & \multicolumn{1}{c|}{\textbf{\#Tokens}} & \multicolumn{1}{c}{\textbf{Time (ms)}} & \multicolumn{1}{c|}{\textbf{\#Tokens}} & \multicolumn{1}{c}{\textbf{Time (ms)}} & \multicolumn{1}{c|}{\textbf{\#Tokens}} & \multicolumn{1}{c}{\textbf{Time (ms)}} & \multicolumn{1}{c}{\textbf{\#Tokens}} & \multicolumn{1}{c}{\textbf{Time/sample (ms)}} & \textbf{Tokens} \\ \midrule
\multirow{10}{*}{\textbf{\rotatebox{90}{Llama-3.1}}} & \multicolumn{1}{c|}{\multirow{5}{*}{\textbf{\rotatebox{90}{8B}}}} & \multicolumn{1}{l|}{\textsc{Base}} & 161.46 & \multicolumn{1}{l|}{6.68} & 290.76 & \multicolumn{1}{l|}{11.79} & 230.18 & \multicolumn{1}{l|}{6.72} & 213.18 & \multicolumn{1}{l|}{10.42} & \multicolumn{1}{l|}{223.90\di{+0.0\%}} & 8.90\di{+0.0\%} \\
 & \multicolumn{1}{c|}{} & \multicolumn{1}{l|}{\textsc{Cot}} & 163.14 & \multicolumn{1}{l|}{6.52} & 290.86 & \multicolumn{1}{l|}{11.77} & 230.21 & \multicolumn{1}{l|}{6.76} & 213.85 & \multicolumn{1}{l|}{10.36} & \multicolumn{1}{l|}{224.51\di{+0.3\%}} & 8.85\di{-0.6\%} \\
 & \multicolumn{1}{c|}{} & \multicolumn{1}{l|}{\textsc{FullElicit}} & 163.55 & \multicolumn{1}{l|}{6.49} & 290.79 & \multicolumn{1}{l|}{11.83} & 236.30 & \multicolumn{1}{l|}{6.97} & 213.64 & \multicolumn{1}{l|}{10.47} & \multicolumn{1}{l|}{226.07\di{+1.0\%}} & 8.94\di{+0.4\%} \\
 & \multicolumn{1}{c|}{} & \multicolumn{1}{l|}{\textsc{PromptElicit}} & 1419.08 & \multicolumn{1}{l|}{75.78} & 1718.02 & \multicolumn{1}{l|}{85.18} & 2412.91 & \multicolumn{1}{l|}{116.81} & 1117.64 & \multicolumn{1}{l|}{61.40} & \multicolumn{1}{l|}{1666.91\di{+644.5\%}} & 84.79\di{+852.5\%} \\
 & \multicolumn{1}{c|}{} & \multicolumn{1}{l|}{\cem{\bf\se}} & \cem204.65 & \multicolumn{1}{l|}{\cem6.98} & \cem318.74 & \multicolumn{1}{l|}{\cem10.54} & \cem297.71 & \multicolumn{1}{l|}{\cem7.44} & \cem233.78 & \multicolumn{1}{l|}{\cem9.83} & \multicolumn{1}{l|}{\cem263.72\di{+17.8\%}} & \cem8.70\di{-2.3\%} \\ \cmidrule(l){2-13} 
 & \multicolumn{1}{c|}{\multirow{5}{*}{\textbf{\rotatebox{90}{70B}}}} & \multicolumn{1}{l|}{\textsc{Base}} & 910.68 & \multicolumn{1}{l|}{5.17} & 1850.04 & \multicolumn{1}{l|}{11.32} & 1397.99 & \multicolumn{1}{l|}{5.95} & 1421.32 & \multicolumn{1}{l|}{11.18} & \multicolumn{1}{l|}{1395.01\di{+0.0\%}} & 8.40\di{+0.0\%} \\
 & \multicolumn{1}{c|}{} & \multicolumn{1}{l|}{\textsc{Cot}} & 917.37 & \multicolumn{1}{l|}{5.15} & 1856.47 & \multicolumn{1}{l|}{10.61} & 1398.79 & \multicolumn{1}{l|}{5.88} & 1423.61 & \multicolumn{1}{l|}{10.88} & \multicolumn{1}{l|}{1399.06\di{+0.3\%}} & 8.13\di{-3.2\%} \\
 & \multicolumn{1}{c|}{} & \multicolumn{1}{l|}{\textsc{FullElicit}} & 915.09 & \multicolumn{1}{l|}{5.41} & 1851.00 & \multicolumn{1}{l|}{11.13} & 1448.56 & \multicolumn{1}{l|}{6.32} & 1434.97 & \multicolumn{1}{l|}{10.31} & \multicolumn{1}{l|}{1412.41\di{+1.2\%}} & 8.29\di{-1.3\%} \\
 & \multicolumn{1}{c|}{} & \multicolumn{1}{l|}{\textsc{PromptElicit}} & 8545.64 & \multicolumn{1}{l|}{69.39} & 7712.69 & \multicolumn{1}{l|}{55.77} & 8283.76 & \multicolumn{1}{l|}{56.14} & 7139.59 & \multicolumn{1}{l|}{60.06} & \multicolumn{1}{l|}{7920.42\di{+467.8\%}} & 60.34\di{+618.1\%} \\
 & \multicolumn{1}{c|}{} & \multicolumn{1}{l|}{\cem{\bf\se}} & \cem1198.53 & \multicolumn{1}{l|}{\cem6.05} & \cem1854.49 & \multicolumn{1}{l|}{\cem9.48} & \cem1672.19 & \multicolumn{1}{l|}{\cem6.15} & \cem1398.81 & \multicolumn{1}{l|}{\cem9.88} & \multicolumn{1}{l|}{\cem1531.00\di{+9.7\%}} & \cem7.89\di{-6.1\%} \\ \midrule
\multirow{10}{*}{\textbf{\rotatebox{90}{Mistral}}} & \multicolumn{1}{c|}{\multirow{5}{*}{\textbf{\rotatebox{90}{7B}}}} & \multicolumn{1}{l|}{\textsc{Base}} & 468.26 & \multicolumn{1}{l|}{21.60} & 564.08 & \multicolumn{1}{l|}{24.60} & 479.97 & \multicolumn{1}{l|}{18.67} & 647.02 & \multicolumn{1}{l|}{31.38} & \multicolumn{1}{l|}{539.83\di{+0.0\%}} & 24.06\di{+0.0\%} \\
 & \multicolumn{1}{c|}{} & \multicolumn{1}{l|}{\textsc{Cot}} & 489.55 & \multicolumn{1}{l|}{23.31} & 568.91 & \multicolumn{1}{l|}{25.27} & 502.71 & \multicolumn{1}{l|}{19.59} & 688.56 & \multicolumn{1}{l|}{34.02} & \multicolumn{1}{l|}{562.43\di{+4.2\%}} & 25.55\di{+6.2\%} \\
 & \multicolumn{1}{c|}{} & \multicolumn{1}{l|}{\textsc{FullElicit}} & 477.01 & \multicolumn{1}{l|}{19.74} & 567.08 & \multicolumn{1}{l|}{23.57} & 479.99 & \multicolumn{1}{l|}{18.61} & 647.23 & \multicolumn{1}{l|}{31.78} & \multicolumn{1}{l|}{542.83\di{+0.6\%}} & 23.42\di{-2.7\%} \\
 & \multicolumn{1}{c|}{} & \multicolumn{1}{l|}{\textsc{PromptElicit}} & 1444.44 & \multicolumn{1}{l|}{70.58} & 2043.11 & \multicolumn{1}{l|}{97.09} & 2230.07 & \multicolumn{1}{l|}{102.03} & 1788.88 & \multicolumn{1}{l|}{89.59} & \multicolumn{1}{l|}{1876.62\di{+247.6\%}} & 89.82\di{+273.3\%} \\
 & \multicolumn{1}{c|}{} & \multicolumn{1}{l|}{\cem{\bf\se}} & \cem330.42 & \multicolumn{1}{l|}{\cem12.74} & \cem493.56 & \multicolumn{1}{l|}{\cem18.33} & \cem440.59 & \multicolumn{1}{l|}{\cem13.32} & \cem462.43 & \multicolumn{1}{l|}{\cem21.00} & \multicolumn{1}{l|}{\cem431.75\di{-20.0\%}} & \cem16.34\di{-32.1\%} \\ \cmidrule(l){2-13} 
 & \multicolumn{1}{c|}{\multirow{5}{*}{\textbf{\rotatebox{90}{12B}}}} & \multicolumn{1}{l|}{\textsc{Base}} & 205.88 & \multicolumn{1}{l|}{5.62} & 380.22 & \multicolumn{1}{l|}{10.46} & 287.35 & \multicolumn{1}{l|}{5.32} & 251.74 & \multicolumn{1}{l|}{7.97} & \multicolumn{1}{l|}{281.30\di{+0.0\%}} & 7.34\di{+0.0\%} \\
 & \multicolumn{1}{c|}{} & \multicolumn{1}{l|}{\textsc{Cot}} & 208.52 & \multicolumn{1}{l|}{5.57} & 381.90 & \multicolumn{1}{l|}{10.09} & 287.36 & \multicolumn{1}{l|}{5.35} & 258.02 & \multicolumn{1}{l|}{7.41} & \multicolumn{1}{l|}{283.95\di{+0.9\%}} & 7.11\di{-3.2\%} \\
 & \multicolumn{1}{c|}{} & \multicolumn{1}{l|}{\textsc{FullElicit}} & 208.39 & \multicolumn{1}{l|}{5.65} & 380.59 & \multicolumn{1}{l|}{10.54} & 289.87 & \multicolumn{1}{l|}{5.41} & 254.23 & \multicolumn{1}{l|}{7.90} & \multicolumn{1}{l|}{283.27\di{+0.7\%}} & 7.37\di{+0.4\%} \\
 & \multicolumn{1}{c|}{} & \multicolumn{1}{l|}{\textsc{PromptElicit}} & 1273.57 & \multicolumn{1}{l|}{46.77} & 1560.04 & \multicolumn{1}{l|}{52.11} & 1749.14 & \multicolumn{1}{l|}{54.64} & 1226.14 & \multicolumn{1}{l|}{46.18} & \multicolumn{1}{l|}{1452.22\di{+416.3\%}} & 49.93\di{+580.0\%} \\
 & \multicolumn{1}{c|}{} & \multicolumn{1}{l|}{\cem{\bf\se}} & \cem278.07 & \multicolumn{1}{l|}{\cem6.29} & \cem428.54 & \multicolumn{1}{l|}{\cem9.02} & \cem385.85 & \multicolumn{1}{l|}{\cem5.45} & \cem260.08 & \multicolumn{1}{l|}{\cem7.47} & \multicolumn{1}{l|}{\cem338.14\di{+20.2\%}} & \cem7.06\di{-3.9\%} \\ \midrule
\multirow{10}{*}{\textbf{\rotatebox{90}{Qwen2.5}}} & \multicolumn{1}{c|}{\multirow{5}{*}{\textbf{\rotatebox{90}{7B}}}} & \multicolumn{1}{l|}{\textsc{Base}} & 143.78 & \multicolumn{1}{l|}{6.36} & 303.72 & \multicolumn{1}{l|}{13.53} & 247.26 & \multicolumn{1}{l|}{8.26} & 286.31 & \multicolumn{1}{l|}{15.49} & \multicolumn{1}{l|}{245.27\di{+0.0\%}} & 10.91\di{+0.0\%} \\
 & \multicolumn{1}{c|}{} & \multicolumn{1}{l|}{\textsc{Cot}} & 449.74 & \multicolumn{1}{l|}{25.10} & 407.30 & \multicolumn{1}{l|}{20.06} & 370.10 & \multicolumn{1}{l|}{15.93} & 460.24 & \multicolumn{1}{l|}{26.51} & \multicolumn{1}{l|}{421.85\di{+72.0\%}} & 21.90\di{+100.7\%} \\
 & \multicolumn{1}{c|}{} & \multicolumn{1}{l|}{\textsc{FullElicit}} & 145.24 & \multicolumn{1}{l|}{6.18} & 306.38 & \multicolumn{1}{l|}{12.89} & 255.94 & \multicolumn{1}{l|}{6.85} & 291.09 & \multicolumn{1}{l|}{16.35} & \multicolumn{1}{l|}{249.66\di{+1.8\%}} & 10.57\di{-3.1\%} \\
 & \multicolumn{1}{c|}{} & \multicolumn{1}{l|}{\textsc{PromptElicit}} & 1109.49 & \multicolumn{1}{l|}{62.77} & 1227.45 & \multicolumn{1}{l|}{64.27} & 1312.63 & \multicolumn{1}{l|}{64.84} & 1005.11 & \multicolumn{1}{l|}{58.32} & \multicolumn{1}{l|}{1163.67\di{+374.4\%}} & 62.55\di{+473.3\%} \\
 & \multicolumn{1}{c|}{} & \multicolumn{1}{l|}{\cem{\bf\se}} & \cem194.21 & \multicolumn{1}{l|}{\cem6.90} & \cem357.74 & \multicolumn{1}{l|}{\cem12.92} & \cem292.41 & \multicolumn{1}{l|}{\cem7.59} & \cem313.18 & \multicolumn{1}{l|}{\cem16.05} & \multicolumn{1}{l|}{\cem289.38\di{+18.0\%}} & \cem10.87\di{-0.4\%} \\ \cmidrule(l){2-13} 
 & \multicolumn{1}{c|}{\multirow{5}{*}{\textbf{\rotatebox{90}{32B}}}} & \multicolumn{1}{l|}{\textsc{Base}} & 624.64 & \multicolumn{1}{l|}{6.55} & 1271.39 & \multicolumn{1}{l|}{15.35} & 773.30 & \multicolumn{1}{l|}{6.51} & 1044.09 & \multicolumn{1}{l|}{14.50} & \multicolumn{1}{l|}{928.35\di{+0.0\%}} & 10.73\di{+0.0\%} \\
 & \multicolumn{1}{c|}{} & \multicolumn{1}{l|}{\textsc{Cot}} & 625.01 & \multicolumn{1}{l|}{7.46} & 1447.84 & \multicolumn{1}{l|}{18.24} & 871.12 & \multicolumn{1}{l|}{8.13} & 1048.22 & \multicolumn{1}{l|}{14.59} & \multicolumn{1}{l|}{998.05\di{+7.5\%}} & 12.10\di{+12.8\%} \\
 & \multicolumn{1}{c|}{} & \multicolumn{1}{l|}{\textsc{FullElicit}} & 636.89 & \multicolumn{1}{l|}{6.63} & 1275.19 & \multicolumn{1}{l|}{15.06} & 780.95 & \multicolumn{1}{l|}{6.80} & 1052.68 & \multicolumn{1}{l|}{14.21} & \multicolumn{1}{l|}{936.43\di{+0.9\%}} & 10.67\di{-0.5\%} \\
 & \multicolumn{1}{c|}{} & \multicolumn{1}{l|}{\textsc{PromptElicit}} & 5556.76 & \multicolumn{1}{l|}{82.86} & 5468.44 & \multicolumn{1}{l|}{76.23} & 5290.17 & \multicolumn{1}{l|}{69.83} & 4096.52 & \multicolumn{1}{l|}{62.62} & \multicolumn{1}{l|}{5102.97\di{+449.7\%}} & 72.89\di{+579.5\%} \\
 & \multicolumn{1}{c|}{} & \cem{\bf\se} & \cem738.98 & \cem6.32 & \cem1220.44 & \cem12.29 & \cem1011.58 & \cem7.02 & \cem948.09 & \cem11.47 & \cem979.77\di{+5.5\%} & \cem9.28\di{-13.5\%} \\ \bottomrule
\end{tabular}%
}
\vspace{-10pt}
\end{table*}

\paragraph{\se implementation details.}
Unless otherwise specified, all \se results in this paper use the default settings stated in the paper: the last 50\% of layers are designated as evidence-reading layers, with an elicitation threshold $\alpha = 0.5$. 
We use SpaCy~\cite{honnibal2017spacy} to segment the context passages into sentences, and \se performs the sentence-level evidence score computation and elicitation based on this segmentation.
Note that this is only for \se, for \textsc{PromptElicit}, we directly search for exact text-level matches of the generatively extracted evidence to ensure that all extracted information is utilized.

\section{Additional Results and Analysis}
\label{sec:apd-res}

\subsection{Computational Efficiency}\label{sec:apd-res-eff}
% \paragraph{Computational efficiency.}
Due to space constraints, we report only the average running time per sample across all datasets in Table~\ref{tab:main}. Table~\ref{tab:efficiency} provides a detailed breakdown of efficiency results for each dataset, including the average inference time per sample (i.e., context-question pair) and the total number of tokens generated during inference.
Note that Mistral-12B generates shorter answers (7.34 tokens on average) for each sample compared to Mistral-7B (24.06 tokens), thus showing a shorter inference time for each sample as it generates fewer tokens.
We observe that: 
(i) While \fe is highly efficient as a naive method, it barely helps with the QA performance, as shown in Table~\ref{tab:main}. 
(ii) \pe, based on generative evidence extraction, requires the model to generate a large volume of extra tokens during inference, up to 852\% more than direct QA, resulting in considerably slower inference times. 
(iii) As discussed before, the response to \co is not consistent for different LMs. Generally, for Llama and Mistral models, \co have little impact on the QA performance, and inference time as well. For Qwen-series models, \co obtains slightly better EM by generating the intermediate reasoning steps before answering, but this also leads to a great drop in F1 score and significantly higher inference time (e.g., +72.0\% for Qwen2.5-7B) due to the generation of intermediate step tokens.
(iv) \se, by helping the model focus on contextual evidence, generates shorter but more accurate answers. Additionally, as \se does not involve generative evidence extraction, it remains computationally efficient, with only 3-5\% of the inference time overhead compared to \pe.

\subsection{More Layer-wise Attention Visualization}\label{sec:apd-res-attlayer}
\begin{figure*}[t]
\vspace{-5pt}
    \centering
    \includegraphics[width=\linewidth]{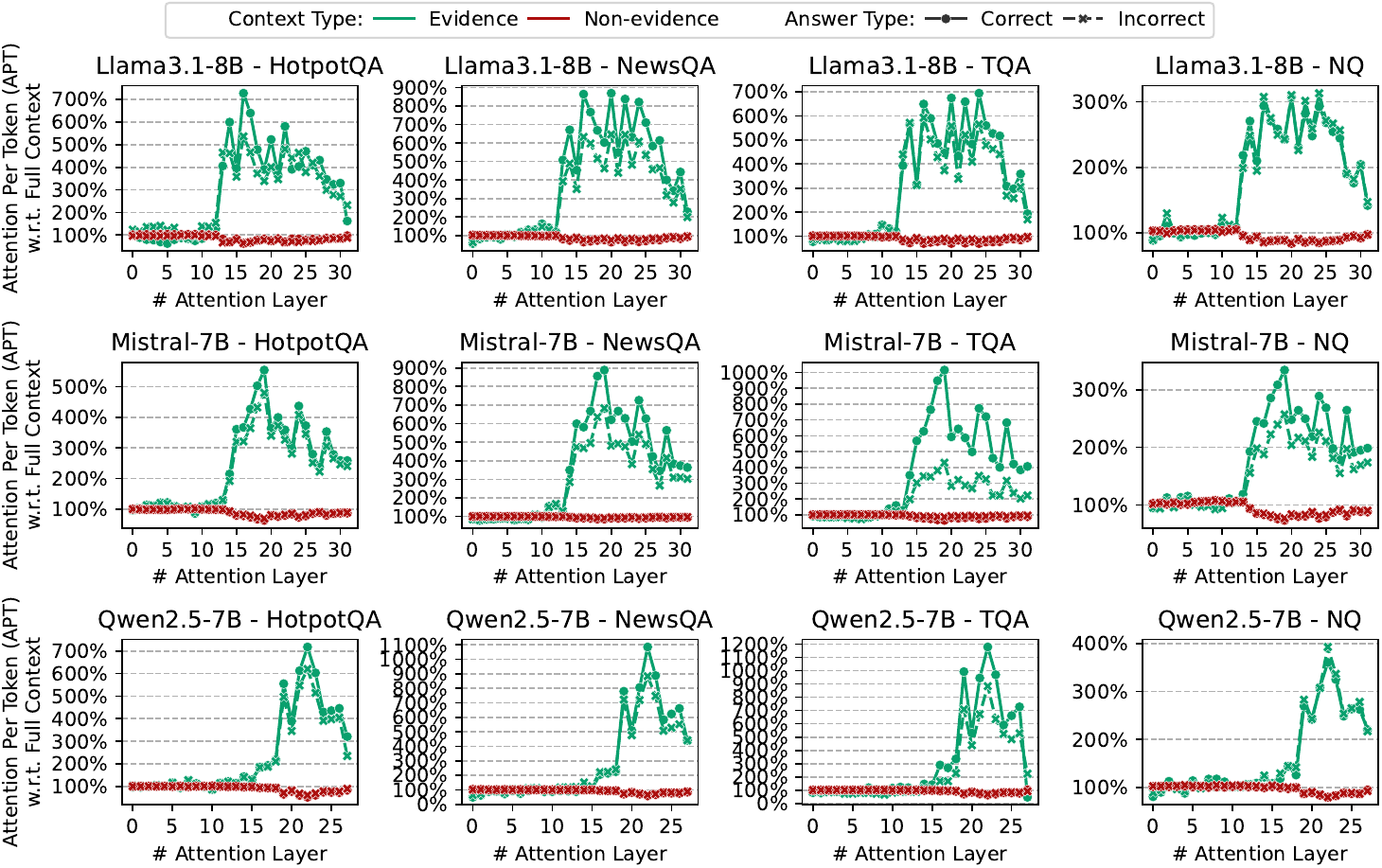}
    \caption[evidence reading layers]{
        Across different LM families and datasets, the deep attention layers highlight the crucial evidence sentences within the context, even when the LM gives incorrect answers (dashed lines).
        The X-axis is the depth of attention layers, and the Y-axis is the ratio of the average attention per token (APT) in the evidence/non-evidence sections to the APT across the entire context.
        For the HotpotQA~\cite{yang2018hotpotqa} dataset, We leverage the ``\texttt{supporting\_facts}'' annotations to differentiate evidence and non-evidence sentences within the context.
        For other datasets, we treat a context sentence as evidence if it contains at least one of the correct answers.
    }
    \label{fig:layeratt-full}
\vspace{-5pt}
\end{figure*}

In Figure~\ref{fig:layeratt}, we present the differences in attention paid by different layers of the model to evidence versus non-evidence context information on the HotpotQA~\cite{yang2018hotpotqa} dataset. We chose HotpotQA because it provides human-annotated sentence-level "supporting\_facts" annotations.
In this section, we extend our analysis to other datasets. Since human annotations are unavailable for these datasets, we use a simple rule to distinguish evidence from non-evidence sentences: any sentence containing at least one correct answer is considered evidence. While this is not a rigorous approach for labeling evidence sentences, it still allows us to visualize patterns and roughly validate whether our observations hold across different datasets.
As shown in Figure~\ref{fig:layeratt-full}, deeper layers of various LMs consistently demonstrate a strong ability to differentiate evidence across datasets.

\subsection{Impact of Evidence Highlighting on Model Attention}
To assess the effectiveness of \se in directing model attention toward relevant context, we analyzed the average attention score per token for the selected evidence sentences before and after highlighting. We conducted this analysis across four QA datasets using Llama-3.1-8B, averaging attention scores over all layers and heads.
As shown in Table~\ref{tab:attention_analysis}, we observe a consistent increase in attention scores for the highlighted evidence across all datasets. This demonstrates that the highlighting strategy effectively shifts model focus toward critical contextual information, reinforcing its ability to retrieve and utilize evidence more effectively.

\begin{table}[h]
\centering
\caption{Effect of highlighting on model attention, measured as average attention per token (APT) for selected evidence sentences.}
\label{tab:attention_analysis}
\vspace{-5pt}
\resizebox{\columnwidth}{!}{%
  \begin{tabular}{lcccc}
    \toprule
    \textbf{APT ($\times 10^{-3}$)} & \textbf{HotpotQA} & \textbf{NewsQA} & \textbf{TQA} & \textbf{NQ} \\ 
    \midrule
    Before Highlighting  & 0.4362  & 0.3465  & 0.2090  & 0.9631  \\ 
    After Highlighting   & 0.6351  & 0.5783  & 0.3451  & 1.1917  \\  
    \midrule
    After/Before Ratio   & 145.61\% & 166.91\% & 165.10\% & 123.73\% \\ 
    \bottomrule
  \end{tabular}
}
\end{table}

\subsection{Sentence-level versus Token-level Eliciting}\label{sec:apd-res-token}
We now show that sentence-level evidence eliciting holds a significant advantage over token-level eliciting due to its ability to highlight evidence with better semantic coherence. 
Table~\ref{tab:sent-vs-token} presents the experimental results using the llama3.1-8B model across four datasets.
We compare the normal version \se (sentence) and a new token-level variant \se (token) by treating each token as a separate "sentence".
It is evident that token-level eliciting performs worse than sentence-level eliciting on all datasets. 
The underlying reason is that tokenization does not ensure each token represents a complete entity or concept, resulting in highlighted evidence that lacks full semantic meaning.
For example, we notice that a year within the context, say "2002", can be tokenized into two tokens "200" and "2" and token-level eliciting could end up highlighting only the first token "200.".
Similar "partial highlighting" issues also occur with names, locations, and other uncommon phrases. 
In contrast, sentence-level eliciting highlights the entire sentence, maintaining semantic continuity in the processed context, which better aids the model in utilizing context information for QA.

\begin{table}[t]
\centering
\caption{Sentence-level versus token-level eliciting.}
\label{tab:sent-vs-token}
\vspace{-5pt}
\resizebox{0.9\columnwidth}{!}{%
\begin{tabular}{@{}ccccc@{}}
\toprule
\textbf{Elicit} & \multicolumn{4}{c}{\textbf{Dataset}} \\ \cmidrule(l){2-5} 
\textbf{Level} & \textbf{HotpotQA} & \textbf{NewsQA} & \textbf{TQA} & \textbf{NQ} \\ \midrule
\multicolumn{5}{c}{Metric: EM} \\ \midrule
\multicolumn{1}{c|}{Token} & 66.5 & 63.5 & 74.0 & 57.5 \\
\multicolumn{1}{c|}{Sentence} & \textbf{68.5} & \textbf{66.9} & \textbf{79.4} & \textbf{64.0} \\ \midrule
\multicolumn{5}{c}{Metric: Token F1} \\ \midrule
\multicolumn{1}{c|}{Token} & 68.6 & 58.4 & 70.9 & 61.6 \\
\multicolumn{1}{c|}{Sentence} & \textbf{69.5} & \textbf{60.8} & \textbf{72.7} & \textbf{67.8} \\ \bottomrule
\end{tabular}%
}
\vspace{-10pt}
\end{table}

\subsection{Alternative Highlighting Strategies}\label{sec:apd-highlighting}

In our primary experiments, we utilized in-context highlighting, where relevant evidence is marked directly within the original passage. This approach aligns with the core insight of \se—leveraging the LM’s inherent ability to identify relevant evidence—while maintaining the semantic continuity of the input. However, alternative methods of evidence integration could potentially influence the model’s ability to utilize highlighted information effectively.
To explore this, we conducted additional experiments using non-in-context highlighting, where all extracted evidence sentences were aggregated and appended either to the beginning or end of the original context. We evaluated these variations using Llama-3.1-8B across four datasets and reported F1 scores in Table~\ref{tab:highlighting}.
\begin{table}[h]
\centering
\caption{Comparison of different highlighting strategies.}
\label{tab:highlighting}
\vspace{-5pt}
\resizebox{\columnwidth}{!}{%
  \begin{tabular}{lcccc}
    \toprule
    \textbf{Highlighting} & \textbf{HotpotQA} & \textbf{NewsQA} & \textbf{TQA} & \textbf{NQ} \\ 
    \midrule
    -        & 57.7  & 56.2  & 66.1  & 61.6  \\ 
    In-context  & \textit{69.5}  & \textbf{60.8}  & \textbf{72.7}  & \textbf{67.8}  \\ 
    Attach at Begin      & \textbf{69.8}  & \textit{59.8}  & \textit{69.3}  & \textbf{67.8}  \\ 
    Attach at End        & 69.0  & 58.8  & 67.0  & 65.6  \\ 
    \bottomrule
  \end{tabular}
}
\vspace{-10pt}
\end{table}
The results indicate that in-context highlighting generally yields the best performance, likely due to its preservation of the original passage’s semantic flow, allowing the model to process highlighted evidence within its natural context. Appending evidence to the beginning or end also improves performance over the base model but is slightly less effective, particularly for multi-hop tasks like HotpotQA, where contextual relationships between facts are crucial.

We emphasize that \se adopts in-context highlighting primarily for its simplicity and efficiency. However, our framework is flexible, and alternative highlighting strategies can be seamlessly incorporated. Exploring more effective context augmentation techniques remains an intriguing direction for future work.

\subsection{Potentially Uninformative Context}
To further analyze \se's robustness in handling cases where the provided context may lack relevant evidence, we constructed an additional QA task based on HotpotQA. This experiment assesses whether \se can correctly reject answers when the context does not support a response. Specifically, we randomly selected 50\% of the samples as unanswerable, replacing their original context with unrelated passages from other samples. The correct response in these cases was set to: "I cannot answer based on the given context."
We explicitly instructed the LLM to generate this response whenever relevant information was absent. In addition to standard EM and F1 scores, we introduced Rejection Accuracy, which measures the model's ability to correctly identify and reject uninformative contexts. The results on Llama-3.1-8B are shown in Table~\ref{tab:notpresent}.

\begin{table}[h]
\centering
\caption{Results on potentially unanswerable QA task.}
\label{tab:notpresent}
\vspace{-5pt}
\resizebox{0.9\columnwidth}{!}{%
  \begin{tabular}{lccc}
    \hline
    \textbf{Method} & \textbf{F1} &\textbf{EM} & \textbf{RejectionAccuracy} \\ 
    \hline
    Base & 78.48 & 79.58 & 91.25 \\ 
    \se & \textbf{82.65} & \textbf{81.93} & \textbf{92.86} \\ 
    \hline
  \end{tabular}
}
\end{table}

These findings, along with Table~\ref{tab:main} in the main paper, demonstrate that \se does not degrade—rather, it slightly enhances—the model’s ability to recognize when the provided context lacks sufficient information for answering. At the same time, \se significantly improves the model’s ability to utilize relevant information when it is available by highlighting the most crucial evidence in relation to the user query.

\subsection{Highlighting versus Filtering Strategies}\label{sec:apd-filco}
In this section, we compare \se to the Filco method which filters context using a fine-tuned Llama-2 model~\cite{wang2023filco}. Additionally, we explore the impact of highlighting versus filtering evidence.
To analyze the impact of different evidence extraction and highlighting strategies, we conduct experiments on four \se/Filco variants:
\textbf{(i) \se-highlight}: The standard \se approach, where identified evidence sentences within the context are explicitly highlighted.
\textbf{(ii) \se-filter}: Instead of highlighting, this method extracts the identified evidence sentences and concatenates them to form a new, filtered context passage.
\textbf{(iii) Filco-filter}: The standard Filco approach, which employs a fine-tuned Llama-2 model to generate a filtered version of the context for answering questions.
\textbf{(iv) Filco-highlight}: This variant highlights the evidence sentences identified by Filco within the original context, rather than filtering them out.

This experimental setup enables a direct comparison between highlighting vs. filtering, as well as \se vs. Filco, in their ability to enhance LMs’ utilization of contextual information. We use Llama-3.1-8B-Instruct as the base LM and evaluate performance across four QA datasets. The F1 scores are presented in Table~\ref{tab:se_filco_comparison}.

\begin{table}[h]
\centering
\caption{Comparison of \se and Filco variants on four QA datasets, evaluated using F1 scores. \se-highlight achieves the highest performance across all datasets.}
\label{tab:se_filco_comparison}
\vspace{-5pt}
\resizebox{\columnwidth}{!}{%
\begin{tabular}{lcccc}
\toprule
\textbf{Method} & \textbf{HotpotQA} & \textbf{NewsQA} & \textbf{TQA} & \textbf{NQ} \\
\midrule
Base & 57.7 & 56.2 & 66.1 & 61.6 \\
\se-highlight & \textbf{69.5} & \textbf{60.8} & \textbf{72.7} & \textbf{67.8} \\
\se-filter & 62.4 & 55.6 & 66.5 & 65.1 \\
Filco-highlight & \textit{67.8} & \textit{57.2} & \textit{68.9} & \textit{65.3} \\
Filco-filter & 62.6 & 54.7 & 66.3 & 64.5 \\
\bottomrule
\end{tabular}
}
\end{table}

The key findings are:
(i) \textbf{\se consistently achieves the highest performance across all datasets, outperforming not only its filtering-based counterpart but also Filco, a more complex, fine-tuned approach.} This highlights the effectiveness of inference-time evidence elicitation over training-based filtering.
(ii) \textbf{Highlighting-based methods outperform their filtering-based counterparts for both \se and Filco.} Filtering removes context that may contain auxiliary or indirect supporting information, such as intermediate reasoning steps, leading to information loss.
(iii) \textbf{Filtering-based methods struggle particularly on NewsQA, which consists of news articles as context.} Unlike structured knowledge sources such as Wikipedia, news articles tend to be semantically cohesive—making it harder to extract self-contained and sufficiently informative evidence sentences. This may explain why Filco was originally evaluated on Wikipedia-based datasets, where evidence extraction is more straightforward.

These findings reinforce that highlighting key evidence within the original context is more effective than filtering, as it preserves both primary evidence and surrounding contextual cues that may contribute to reasoning. Moreover, \se, despite being a lightweight inference-time method, surpasses Filco, which relies on additional model fine-tuning—demonstrating the advantage of leveraging pretrained LMs’ inherent ability to locate and utilize relevant evidence.

\subsection{Examples with full context}\label{sec:apd-res-exa}
% \paragraph{Examples with full context.}
Finally, we present the full context of examples from Table~\ref{tab:example} to illustrate \se's ability to identify supporting facts/evidence within noisy long contexts. The results are shown in Tables~\ref{tab:example-full1} and \ref{tab:example-full2}.
These examples are from HotpotQA's distractor setting, where the context passage contains distracting information retrieved from Wikipedia using the question as a query~\cite{yang2018hotpotqa}.
Note that these distractors are not completely irrelevant random noise that can be easily filtered out by retrieval systems. 
Instead, they appear related to the question but do not actually support answering it, serving as "hard negatives."
This scenario is quite common in practice and significantly impacts the performance of retrieval-augmented generation~\cite{wu2024ragirrelevant,cuconasu2024ragnoise}.
As shown in Tables~\ref{tab:example-full1} and \ref{tab:example-full2}, the LM struggles to effectively use contextual evidence to provide correct answers under such a situation. 
Despite this, \se accurately highlights the critical evidence within the noisy long context, helping the model focus on the most relevant facts and thus arrive at the correct answer.

\section{Additional Discussions}

\subsection{Failure Case Analysis and Potential Improvements}\label{sec:apd-failcase}

While \se significantly enhances LM’s ability to identify and utilize relevant evidence, there are cases where it fails to aid the model in generating correct answers. Through manual analysis, we found that incomplete evidence highlighting is a primary cause of failure.
Consider the following example where Llama-3.1-8B, both with and without \se, gives a negative response.
\begin{tcolorbox}[title={An example case with $\alpha=0.5$},top=1mm,bottom=1mm]
\scriptsize
\textbf{Context:} Alexander S. Wadsworth Commodore Alexander Scammel Wadsworth (1790–April 5, 1851) was an officer of the United States Navy. His more than 40 years of active duty included service in the War of 1812. War of 1812 \textev{The War of 1812 (1812–1815) was a conflict fought between the United States, the United Kingdom, and their respective allies.} Historians in Britain often see it as a minor theater of the Napoleonic Wars; in the United States and Canada, it is seen as a war in its own right ...  \\
\textbf{Question:} Alexander S. Wadsworth served during the war that was fought between which two countries?  \\
% \textbf{Correct Answer:} United States, the United Kingdom, and their respective allies.  \\
\textbf{Base Model:} I cannot answer based on the given context. \xmark \\
\textbf{\se:} I cannot answer based on the given context. \xmark 
\end{tcolorbox}
We can see that although \se successfully highlighted the sentence: "The War of 1812 (1812–1815) was a conflict fought between the United States, the United Kingdom, and their respective allies." as important evidence, this alone was insufficient for the model to infer that Alexander S. Wadsworth served in this war. 
We observed that lowering $\alpha$ can mitigate such failures by allowing \se to highlight more comprehensive evidence. Reducing $\alpha$ to 0.1 enabled the method to additionally highlight the missing link—“Commodore Alexander Scammel Wadsworth (1790–April 5, 1851) was an officer of the United States Navy. His more than 40 years of active duty included service in the War of 1812.”, and lead to the correct answer.
\begin{tcolorbox}[title={An example case with $\alpha=0.1$},top=1mm,bottom=1mm]
\scriptsize
\textbf{Context:} Alexander S. Wadsworth \textev{Commodore Alexander Scammel Wadsworth (1790–April 5, 1851) was an officer of the United States Navy. His more than 40 years of active duty included service in the War of 1812.} War of 1812 \textev{The War of 1812 (1812–1815) was a conflict fought between the United States, the United Kingdom, and their respective allies.} Historians in Britain often see it as a minor theater of the Napoleonic Wars; in the United States and Canada, it is seen as a war in its own right ...  \\
\textbf{Question:} Alexander S. Wadsworth served during the war that was fought between which two countries?  \\
\textbf{Base Model:} I cannot answer based on the given context. \xmark \\
\textbf{\se:} United States, the United Kingdom, and their respective allies. \cmark 
\end{tcolorbox}

This suggests that a fixed threshold may not be optimal across different questions and contexts, and \se could potentially benefit from a more adaptive thresholding mechanism: Instead of using a fixed $\alpha$, a dynamic threshold that adjusts based on the complexity of the input context or question could provide more flexible evidence selection.
Nevertheless, as shown and discussed in Figure~\ref{fig:threshold}, \se remains robust to the choice of $\alpha$, with a broad range of values yielding consistently strong improvements. Even with static thresholding, \se provides stable and significant gains in LM performance across QA tasks.

\subsection{Understanding Layer-Wise Evidence Utilization in Language Models}

Our analysis reveals that while intermediate layers of language models effectively emphasize relevant evidence within the context, this emphasis does not necessarily translate into correct final responses. As shown in Figure~\ref{fig:layeratt} and \ref{fig:layeratt-full}, attention to key evidence tends to diminish from mid-layers to the final layers, indicating that the final layers do not prioritize contextual evidence as strongly as mid-layers do. This suggests that although models inherently recognize relevant information at intermediate layers, their final token selection during decoding is influenced by additional factors that may override this evidence, leading to incorrect responses.

This phenomenon aligns with recent findings in closed-book QA and controlled generation settings. Prior work has shown that factually correct tokens often receive higher decoding probabilities in intermediate layers, while incorrect tokens can maintain consistently high probabilities across all layers and still be selected as the final output~\cite{chuang2023dola}. Similarly, studies on the "overthinking" behavior of LMs indicate that removing the last few layers can sometimes improve factual accuracy, implying that deeper layers may introduce unnecessary complexity or spurious reasoning \cite{halawi2024overthinking}. Our findings provide further empirical support for these observations, explaining why models can correctly identify relevant evidence yet still produce incorrect answers.

\subsection{Potential Risks}

Generative AI tools such as language models have an increasing impact on our daily lives in the era of big data and AI~\cite{yan2021dynamic,yan2022dissecting,yan2023trainable,yan2023reconciling,yan2024thegcn,yan2024pacer,yan2024topological,xuslog,xu2024discrete,ban2021ee,lin2024bemap,lin2024backtime,guo2023taming,wei2025webagent,wei2024robust,qiu2024efficient,qiu2024ask,qiu2024tucket,qiu2023reconstructing,qiu2022dimes,zeng2023parrot,zeng2023generative,zeng2024hierarchical,zeng2024graph,qiu2024gradient,weiconnecting,ai2023mlp,yoo2025embracing,yoo2025generalizable,yoo2024ensuring,he2023robust,bao2023adaptive,wei2023ntk,wei2024towards,wei2022comprehensive,wei2022augmentations,wei2020fast,wei2021model,li2025zero}, such as finance~\cite{chan2024group,liu2024aim} and healthcare~\cite{ye2023web,liu2024class}, especially with the recent trends of foundation models \cite{DBLP:conf/kdd/ZhengJLTH24, DBLP:journals/corr/abs-2410-12126, DBLP:journals/corr/abs-2412-21151, DBLP:journals/corr/abs-2412-08174,zhang2025improving,ai2025resmoe}.
This study focuses on enhancing language models' ability to effectively utilize information from contextual documents. However, retrieved documents from the internet may contain unethical or discriminatory content, which the model might read and incorporate into its outputs. While addressing ethical concerns in Retrieval-Augmented Generation (RAG) or fact-based question-answering tasks~\cite{cornnet,binet,kompare} is beyond the scope of this work, such issues can typically be mitigated by using detectors to filter harmful information from the context documents provided to the language model, or other general techniques such as ensemble multi-model answers~\cite{liu2021imbens,liu2020self,liu2020mesa} or watermarking~\cite{chen2024wapiti} AI-generated contents.

\subsection{Usage of Artifacts and AI Assistants}

All models and datasets used in this study are publicly available on HuggingFace, and we adhered to their respective licenses and terms of use, limiting our work to non-commercial academic research. These models and datasets have been reviewed by their developers/creators to minimize the inclusion of personally identifiable information or offensive content and are widely adopted by the research community. The datasets primarily consist of English-language content and focus on fact-based question-answering tasks. We used AI tools to assist with language refinement during the writing process, but the paper contains no AI-generated paragraphs. All material has been carefully reviewed to ensure accuracy and adherence to ethical standards.

\begin{table*}[t]
\scriptsize
\centering
\caption{
Full context of example \#1 \& \#2 from Table~\ref{tab:example} demonstrating how our method assists the LM in context-based QA, with \textev{blue} text highlighting the evidence selected by \se. The base LM used is Llama3.1-8B-Instruct.
}
\label{tab:example-full1}
\vspace{-5pt}
\resizebox{\textwidth}{!}{%
\begin{tabular}{p{0.01\textwidth} p{0.75\textwidth} p{0.22\textwidth}}
\toprule
{\bf\small} & {\bf\small Full Context Passage} & {\bf\small Question \& Answers}
\\ \midrule
\multirow{7}{*}{\rotatebox{90}{\bf True or False}}
& 
\textev{Tiger Please is an Indie / Alternative five-piece band from Cardiff, Wales. The band formed in August 2008.} The band's influences are U2, Sigur Rós, Kings of Leon, John Mayer and Counting Crows. They signed with Walnut Tree Records in 2009 and released their debut EP "They Don't Change Under Moonlight". "Kerrang!" magazine, "Rock Sound" magazine, and "Classic Rock" magazine praised the EP and featured the band on the "Rock Sound" and "Classic Rock" cover-mount albums. The band toured with Kids In Glass Houses, InMe, Twin Atlantic and Funeral For A Friend. Taxi Violence is a South African rock band from Cape Town. The group consists of George van der Spy (vocals), Jason Ling (bass), Louis Nel (drums), Rian Zietsman (guitar) and Loedi van Renen (guitar/bass). They have released five studio albums: "Untie Yourself" (2006), "The Turn" (2009), "" (2011), "Soul Shake" (2013), and "Tenfold" (2014). They are influenced by bands such as: Black Rebel Motorcycle Club, Queens of the Stone Age, Led Zeppelin, Jimi Hendrix, Nirvana, and Pearl Jam. This is discography of the American rock band Black Rebel Motorcycle Club \textev{Black Rebel Motorcycle Club (often abbreviated as BRMC) is an American rock band from San Francisco, California.} The group consists of Peter Hayes (vocal, guitar, harmonica), Robert Levon Been (vocal, bass, guitar), and Leah Shapiro (drums). Former drummer Nick Jago left the band in 2008 to focus on his solo project. Skybombers is a rock band from Melbourne. They were formed as Collusion by Scotch College students Hugh Gurney, Ravi Sharma, Scott McMurtrie and Sam Bethune. They later changed to Skybombers, a name inspired by an icy-pole. Their placing a demo song "It Goes Off" on MySpace brought them their first TV appearances. They had early international attention when "It Goes Off" of their EP "Sirens" made the most-played list on L.A.'s Indie 103.1 and played a showcase gig at The Viper Room. They have toured Australia, Japan and USA. and their debut album "Take Me To Town" was recorded in L.A. with Rick Parker (Black Rebel Motorcycle Club). The band made their way on video game media in 2007 when "It Goes Off" appeared on the soundtrack for "Burnout Dominator", the song later reappeared on "Burnout Paradise" in early 2008. Black Carousel was recorded in LA, again with Rick Parker at the helm. Masaki Liu, sometimes referred to as "Saki", is the engineer and producer operating One Way Studio, a digital recording studio in Benicia, California. Masaki has recorded and produced music for many bands, including Five Iron Frenzy, Black Rebel Motorcycle Club, The Echoing Green, The W's and Yellow Second. Dan Russell is a musician and songwriter in addition to an artist manager and advocate, musician, songwriter, concert promoter, record producer and music supervisor for television and film. A graduate of Walpole High School in Massachusetts and later Barrington College, Russell is known for managing both the American rock band The Call and songwriter Michael Been and has worked in various capacities with such artists as Black Rebel Motorcycle Club, Sam Philips, Mark Heard, U2 and Robin Lane, Ramona Silver, Vigilantes of Love, among others. Black Rebel Motorcycle Club: Live is a DVDs of Black Rebel Motorcycle Club concert footage captured from three sold out shows in Berlin, Dublin and Glasgow, and chronicles the end of the band's 2007 tour in support of "Baby 81". Additionally, it includes intimate, behind-the-scenes footage, glimpses into the making of 2005's Howl and is rounded out with a bonus live album featuring 14 songs. KAV (Kav Sandhu) is a British musician from Leicester UK, based in Los Angeles. KAV played guitar with British band Happy Mondays for 4 years after helping reform the band with frontman Shaun Ryder in 2004. He launched his solo project under moniker "KAV" in 2008 with long-time friend and drummer Jim (James) Portas. His solo material has been compared by the media to everyone from Iggy \& The Stooges, Black Rebel Motorcycle Club, Primal Scream, Kasabian, The Rolling Stones \& Bob Dylan. He plays live with a full live band, which sometimes features guest musicians from various bands. Vagrant Records is an American record label based in California. It was founded in 1995 by Rich Egan and Jon Cohen. The label focuses on rock but features artists in a variety of other genres including folk, soul, electronic, and pop. It is home to artists such as The 1975, Death Spells, Eels, Bad Suns, Edward Sharpe and the Magnetic Zeroes, CRUISR, Active Child, PJ Harvey, School of Seven Bells, Black Rebel Motorcycle Club, James Vincent McMorrow, Black Joe Lewis, Wake Owl, Blitzen Trapper, and Bombay Bicycle Club. Originally, Vagrant Records was mostly focused on emo bands such as Dashboard Confessional, Saves the Day, The Get Up Kids, and Alkaline Trio.
& 
{\scriptsize
    \textbf{Question:} Are the bands Tiger Please and Black Rebel Motorcycle Club from the same country?\vspace{1mm}
    
    \textbf{True Answer:} No. \vspace{1mm}
    
    \uline{\textbf{\texttt{Base:}}}
    Yes. \xmark \vspace{1mm}
    
    \uline{\textbf{\texttt{+\se:}}}
    No. \cmark \vspace{1mm}
}
\\ \midrule
\multirow{8}{*}{\rotatebox{90}{\bf Comparison}}
& 
% \uline{\textbf{(Displaying 146 of 532 Words)}}
\textev{Home Monthly was a monthly women's magazine published in Pittsburgh, Pennsylvania in the late 19th century.}
"The Strategy of the Were-Wolf Dog" is a short story by Willa Cather. \textev{It was first published in "Home Monthly" in December 1896.}
The Count of Crow's Nest is a short story by Willa Cather. It was first published in "Home Monthly" in October 1896.
\textev{Mirabella was a women's magazine published from June 1989 to April 2000.} It was created by and named for Grace Mirabella, a former "Vogue" editor in chief, in partnership with Rupert Murdoch.
"Nanette: An Aside" is a short story by Willa Cather. It was first published in "Courier" on 31 July 1897 and one month later in "Home Monthly".
"The Prodigies" is a short story by Willa Cather. It was first published in "Home Monthly" in July 1897.
"A Resurrection" is a short story by American writer Willa Cather. It was first published in "Home Monthly" in April 1897.
Tommy, the Unsentimental is a short story by Willa Cather. It was first published in "Home Monthly" in August 1896.
The Princess Baladina is a short story by Willa Cather. It was first published in "Home Monthly" in 1896 under the pseudonym of Charles Douglass.
"The Way of the World" is a short story by Willa Cather. It was first published in "Home Monthly" in April 1898.
& 
{\scriptsize
    \textbf{Question:} Which women's magazine was published first, Mirabella or Home Monthly?\vspace{1mm}
    
    \textbf{True Answer:} Home Monthly. \vspace{1mm}
    
    \uline{\textbf{\texttt{Base:}}}
    Mirabella. \xmark \vspace{1mm}
    
    \uline{\textbf{\texttt{+\se:}}}
    Home Monthly. \cmark \vspace{1mm}
}
\\
\bottomrule
\end{tabular}
            }
\vspace{-10pt}
\end{table*}

\begin{table*}[t]
\scriptsize
\centering
\caption{
Full context of example \#3 \& \#4 from Table~\ref{tab:example} demonstrating how our method assists the LM in context-based QA, with \textev{blue} text highlighting the evidence selected by \se. The base LM used is Llama3.1-8B-Instruct.
}
\label{tab:example-full2}
\vspace{-5pt}
\resizebox{\textwidth}{!}{%
\begin{tabular}{p{0.01\textwidth} p{0.75\textwidth} p{0.22\textwidth}}
\toprule
{\bf\small} & {\bf\small Full Context Passage} & {\bf\small Question \& Answers}
\\ \midrule
\multirow{7}{*}{\rotatebox{90}{\bf Fact Retrieval}}
& 
The 1987 European Cup Final was a football match held at the Prater Stadium, Vienna, on 27 May 1987, that saw Porto of Portugal defeat Bayern Munich of West Germany 2–1. Both sides were missing key players: the Portuguese were without their injured striker Fernando Gomes, while the Germans were missing their sweeper, and captain, Klaus Augenthaler, who was suspended, along with striker Roland Wohlfarth and midfield player Hans Dorfner, who were both injured. The Portuguese side fought back from 1–0 down to win their first European Cup, with the goals coming from a back heel by Rabah Madjer and a volley from Juary, after a Ludwig Kögl header had given Bayern the lead in the first half. The final was the first European Cup final that Bayern, and their captain Lothar Matthäus would lose to successive late goals, repeated 12 years later in the 1999 UEFA Champions League Final against Manchester United.
The 1993 European Cup Winners' Cup Final was a football match contested between Parma of Italy and Royal Antwerp of Belgium. The final was held at Wembley Stadium in London, England on 12 May 1993. It was the final match of the 1992–93 European Cup Winners' Cup and the 33rd European Cup Winners' Cup Final. Parma beat Antwerp 3–1 and in doing so became the eighth different Italian team to win a European trophy.
The 1970 European Cup Winners' Cup Final was a football match between Manchester City of England and Górnik Zabrze of Poland on 29 April 1970 at Prater Stadium in Vienna, Austria. It was the final match of the 1969–70 European Cup Winners' Cup and the tenth European Cup Winners' Cup final. Both sides made their first appearance in a European final. Manchester City won the match 2–1 thanks to goals by Neil Young and Francis Lee. The victory was City's only European trophy.
The 1977 European Cup Final was an association football match between Liverpool of England and Borussia Mönchengladbach of Germany on 25 May 1977 at the Stadio Olimpico in Rome, Italy (the venue was decided in Bern by the UEFA Executive Committee on 17 September 1976). The showpiece event was the final match of the 1976–77 season of Europe's premier cup competition, the European Cup. Both teams were appearing in their first European Cup final, although the two sides had previously met in the 1973 UEFA Cup Final, which Liverpool won 3–2 on aggregate over two legs.
The 1988 European Cup Winners' Cup Final was a football match contested between Mechelen of Belgium and the defending champions, Ajax of Netherlands. It was the final match of the 1987–88 European Cup Winners' Cup and the 28th European Cup Winners' Cup Final. The final was held at Stade de la Meinau in Strasbourg, France. Mechelen won the match 1–0 thanks to a goal by Piet den Boer.
The 1961 European Cup Winners' Cup Final was a football match contested between Fiorentina of Italy and Rangers of Scotland. It was the final of the 1960–61 European Cup Winners' Cup the first UEFA Cup Winners' Cup final. It was the only time that the final was played over two legs. The first leg was played at Ibrox Stadium, Glasgow and the second leg at the Stadio Comunale in Florence. It was Rangers first European final and in doing so became the first British team to reach the final of a European football competition. It was Fiorentina's second European final having previously reached the 1957 European Cup final.
The 1987 European Cup Winners' Cup Final was a football match contested between Ajax of Netherlands and Lokomotive Leipzig of East Germany. It was the final match of the 1986–87 European Cup Winners' Cup and the 27th European Cup Winners' Cup Final. The final was held at Olympic Stadium in Athens, Greece. Ajax won the match 1–0 with a 20th-minute header from Marco van Basten.
Lars Lunde (born 21 March 1964) is a Danish former professional football player, who played in the striker position. Lunde got his breakthrough with Brøndby IF in 1983, and he made his debut for the Denmark national football team in October 1983. He was sold to Young Boys Bern in Switzerland, before moving to German club Bayern Munich in 1986. \textev{He was a part of the Bayern team which won the German Bundesliga championship in 1987, and he came on as a late substitute when Bayern lost the 1987 European Cup Final to FC Porto.} He played the last of his three matches for the Danish national team in April 1987, before leaving Bayern during the 1987–88 season. He went on to play for a number of smaller clubs, ending his career with FC Baden in Switzerland.
The 1978 European Cup Final was an association football match between Liverpool of England and Club Brugge of Belgium on 10 May 1978 at Wembley Stadium, London, England (the venue was decided in Bern by the UEFA Executive Committee on 20 September 1977). It was the final match of the 1977–78 season of Europe's premier cup competition, the European Cup. Liverpool were the reigning champions and were appearing in their second European Cup final. Club Brugge were appearing in their first European Cup final. The two sides had met once before in European competition, when they contested the 1976 UEFA Cup Final, which Liverpool won 4–3 on aggregate.
The 1985 European Cup Winners' Cup Final was a football match contested between Everton of England and Rapid Wien of Austria. It was the final match of the 1984–85 European Cup Winners' Cup and the 25th European Cup Winners' Cup Final. The final was held at Feijenoord Stadion in Rotterdam, Netherlands, on 15 May 1985. Everton, which dominated throughout, won the match 3–1 thanks to goals by Andy Gray, Trevor Steven and Kevin Sheedy. Everton were unable to defend the trophy: as league champions they would have entered the 1985–86 European Cup, but they were not permitted to play in either competition following the actions of rival Liverpool fans at the Heysel Stadium, which saw all English clubs banned from European competitions.
& 
{\scriptsize
    \textbf{Question:} Which team did Lars Lunde play for when defeated for the 1987 European Cup Final?\vspace{1mm}
    
    \textbf{True Answer:} Bayern Munich \vspace{1mm}
    
    \uline{\textbf{\texttt{BASE:}}}
    FC Porto. \xmark \vspace{1mm}
    
    \uline{\textbf{\texttt{+\se:}}}
    Bayern Munich. \cmark \vspace{1mm}
}
\\ \midrule
\multirow{8}{*}{\rotatebox{90}{\bf Multi-hop Reasoning}}
& 
Mount Barker Junction railway station is a disused station on the Adelaide to Wolseley line serving the South Australian city of Mount Barker. Mount Gambier railway station was the junction station for the Naracoorte–Millicent and Mount Gambier-Heywood lines in the South Australian city of Mount Gambier. Frewville is a small suburb in the South Australian city of Adelaide. It is three kilometres south-east of Adelaide's central business district (CBD). The 2006 South Australian Super League was the first season of the South Australian Super League, the new top division of association football in South Australia, replacing the South Australian Premier League, which became the second division. It was also the first year that football in South Australia was run by the Football Federation of South Australia, which replaced the South Australian Soccer Federation. The season came down to a final round relegation battle between White City Woodville and Adelaide Olympic. Olympic lost 3–1 at Modbury while White City went down 1–0 away to Cumberland. This sent Olympic down to play in the Premier League in 2007. Adelaide City won the title with games to spare after being runaway leaders, finishing the season unbeaten. \textev{Norwood is a suburb of Adelaide, about 4 km east of the Adelaide city centre.} The suburb is in the City of Norwood Payneham \& St Peters, the oldest South Australian local government municipality, with a city population over 34,000. Whyalla railway station was the terminus station of the Whyalla line serving the South Australian city of Whyalla. \textev{Walter Frank Giffen (20 September 1861 in Norwood – 28 June 1949 in Adelaide) was an Australian cricketer who played in 3 Tests between 1887 and 1892.} He was the brother of the great all-rounder George Giffen. The City of Burnside is a local government area with an estimated population of 44,300 people in the South Australian city of Adelaide. Burnside was founded in August 1856 as the District Council of Burnside, and was classed as a city in 1943. It is named after the property of an early settler and stretches from the Adelaide Parklands into the Adelaide foothills. It is bounded by Adelaide, Adelaide Hills Council, Campbelltown, Mitcham, Norwood Payneham and St Peters and Unley. The city has an area of 27.53 km². Glenunga is a small southern suburb of 2,539 people in the South Australian city of Adelaide. It is located five kilometres southeast of the Adelaide city centre. The name Glenunga is taken from an Aboriginal language "unga" meaning near and "glen" because of its proximity to Glen Osmond (see Manning's places of South Australia by Geoffrey H. Manning published in 1990). Bounded on the north by Windsor Road, the east by Portrush Road, the south-west by Glen Osmond Road and the west by Conyngham Street, the leafy suburb forms a rough triangular layout. It is close by to other Burnside council suburbs of Toorak Gardens and Glenside. Collina is a suburb of the Australian city of Griffith in the Riverina region of New South Wales. The suburb is in the City of Griffith local government area. Collina is 4 km northwest of the Griffith city centre and reflects the city's rapid growth in the early 2000s.
& 
{\scriptsize
    \textbf{Question:} Walter Giffen is from a suburb of which South Australian city?\vspace{1mm}
    
    \textbf{True Answer:} Adelaide\vspace{1mm}
    
    \uline{\textbf{\texttt{Base:}}}
    Norwood. \xmark \vspace{1mm}
    
    \uline{\textbf{\texttt{+\se:}}}
    Adelaide. \cmark \vspace{1mm}
}
\\
\bottomrule
\end{tabular}
}
\vspace{-10pt}
\end{table*}

\end{document}